\documentclass{bmvc2k}

\usepackage{wrapfig}
\usepackage{graphicx}
\usepackage{booktabs}
\usepackage{comment}
\usepackage{algorithm}
\usepackage{algorithmic}
\usepackage{multirow}
\usepackage{xcolor}

\usepackage{tabularx}
\usepackage{cleveref}
\usepackage{colortbl}
\usepackage{orcidlink}
\definecolor{lime}{rgb}{0.88,2,10}

\usepackage{enumitem}
\usepackage[spaces,hyphens]{xurl}
\usepackage{amssymb}
\usepackage{textcomp}

\usepackage[accsupp]{axessibility}

\newcommand\HUGE{\fontsize{17}{25}\selectfont}

\definecolor{lightred}{rgb}{1.0, 0.9, 0.9}

\title{\HUGE{Beyond Small Patches: Black-Box Detection and Purification of Diverse Backdoor Triggers}}

\addauthor{Ahmed Abdelnaby}{ahmed.abdelnaby@wsu.edu}{1}
\addauthor{Mohamed Elmahallawy}{mohamed.elmahallawy@wsu.edu}{1}

\addinstitution{
School of Engineering and Applied Sciences, Washington State University, Richland, WA 99354, USA
}

\runninghead{Ahmed Abdelnaby and Mohamed ELmahallawy}{Beyond Small Patches}

\begin{document}

\maketitle
 

\begin{abstract}
Deep neural networks (DNNs) are increasingly deployed in real-world vision systems, yet their predictions can be covertly manipulated by backdoor attacks, in which malicious triggers cause targeted misclassification while preserving high clean accuracy. Existing defenses often rely on model internals, training data, or clean validation samples, making them difficult to deploy when only black-box access to a trained model is available. We propose \textit{TRIM} (\textit{Trigger Removal by Identifying Manipulated Regions}), a deployment-oriented black-box defense that detects and selectively removes backdoor triggers at {\em inference time without requiring model internals, training data, or clean samples}. The key insight behind TRIM is to identify image regions that are responsible for anomalous model behavior and purify only those regions while preserving benign content. TRIM innovates via three key components: (i) region-based segmentation with deep feature representations, (ii) adaptive trigger discovery through inpainting and diffusion-based reconstruction to isolate regions responsible for misclassification---without assumptions about trigger type, shape, or location, and (iii) selective region purification that cleans poisoned regions while retaining benign content. To support practical deployment, TRIM further caches feature embeddings of previously identified triggers, enabling efficient recognition and avoiding redundant detection and purification. Extensive experiments across diverse datasets and backdoor types, including blended, sparse, varying-size, and multiple triggers, show that TRIM consistently outperforms existing black-box defenses, reducing attack success rates (ASR) to as low as \textit{1.16\%} while preserving clean accuracy of up to \textit{87.87\%}. These results demonstrate that effective backdoor mitigation is possible at inference time even when the defender has no access to any auxiliary data\footnote{Our code is publicly available at: \href{https://github.com/wsu-cyber-security-lab-ai/TRIM.git}{GitHub Repository}.}.

\end{abstract}

\vspace{-0.53cm}
\section{Introduction}\label{sec:intro}

Deep Neural Networks (DNNs) are widely deployed in computer vision systems but remain vulnerable to \emph{backdoor attacks}, which activate attacker-defined behavior when specific triggers appear~\cite{gu2017badnets,chen2017targeted,li2022backdoor}. Backdoored models typically retain high accuracy on benign inputs while producing malicious predictions on triggered samples~\cite{liu2018trojaning,gu2017badnets}, making them difficult to detect through conventional evaluation.

Modern triggers extend beyond fixed patches to blended, distributed, dynamic, and sample-specific patterns. Dynamic and input-aware attacks further vary trigger appearance or location, weakening defenses that assume predefined trigger characteristics. Such vulnerabilities can originate from poisoned datasets, \emph{untrusted} pre-trained models, or compromised training pipelines~\cite{bagdasaryan2020backdoor,doan2021backdoor}. Practical defenses should therefore operate without prior knowledge of trigger appearance, size, or location.

Existing \emph{white-box} defenses often require training data, model parameters, internal activations, or trusted validation samples~\cite{zhang2023backdoor,guan2024backdoor,chen2025refine,hou2025flare}, which may be unavailable for models accessed only through prediction interfaces \cite{abdelnaby2026region}. Existing \emph{black-box} defenses~\cite{shi2023black,gao2020backdoor} may instead transform predefined regions, potentially damaging benign content while missing subtle or adaptive triggers. The key challenge is therefore to identify and neutralize trigger-related regions using only black-box model feedback.

We propose \textit{TRIM} (\textit{Trigger Removal by Identifying Manipulated Regions}), a model-agnostic, inference-time black-box defense requiring no training data, model internals, clean validation samples, or trigger knowledge. \textit{TRIM} segments the input, extracts region-level features, and filters candidates using semantic constraints and cached benign/malicious representations. Candidate regions are selectively reconstructed and queried against the classifier; regions whose suppression influences the prediction are localized and jointly purified while preserving benign content. Feature caching reduces redundant verification, while modular segmentation and inpainting provide a tunable robustness--efficiency trade-off. In summary, our main contributions are:
\begin{itemize}[leftmargin=*]\vspace{-1mm}
    \item We introduce \textit{TRIM}, a model-agnostic black-box defense that detects and removes backdoor triggers without access to model internals, training data, clean validation samples, or prior trigger knowledge.
    
   \item We develop a region-level trigger localization mechanism that combines semantic filtering, feature-based candidate selection, and selective reconstruction to identify prediction-critical manipulated regions while preserving benign image content.


    \item We design a feature-caching mechanism that reuses previously verified benign and suspicious representations, reducing redundant model queries and improving the efficiency of inference-time defense.
    

 
    \item We conduct extensive experiments across blended, dynamic, distributed, input-aware, and multi-trigger attacks, demonstrating that \textit{TRIM} reduces attack success rate to as low as $1.16\%$ while maintaining competitive clean accuracy up to $87.87\%$, outperforming state-of-the-art defenses.
\end{itemize}

\section{Related Work}\label{sec:rel_work}

\noindent{\bf Backdoor Attacks in DNNs.}
Backdoor attacks compromise DNNs by embedding malicious behaviors through
poisoned training data or manipulated model parameters while preserving benign
performance. Representative attacks include BadNets~\cite{gu2017badnets},
label-consistent attacks~\cite{turner2019label}, and simulated physical-patch
attacks~\cite{wenger2021backdoor}. Their objectives are commonly categorized
as \emph{All-to-One (A2O)}, which redirects triggered inputs to a single target;
\emph{All-to-All (A2A)}, which maps source classes to predefined incorrect
classes~\cite{gu2017badnets}; and \emph{untargeted (UT)}, which induces
arbitrary incorrect predictions~\cite{li2022backdoor}.

Triggers vary in spatial coverage~\cite{turner2019label,barni2019new}, fusion
strategy~\cite{chen2017targeted,wang2021backdoor}, and whether they are
sample-specific or sample-agnostic~\cite{li2020invisible,gu2017badnets}.
Attacks may also be label-consistent~\cite{turner2019label} or
label-inconsistent~\cite{bai2024backdoor}. Recent dynamic, distributed, and
input-dependent attacks further vary trigger appearance or location across
samples, while reactivation attacks can restore dormant backdoors through
minor input modifications~\cite{zhu2024breaking}. This diversity motivates
defenses without fixed assumptions about trigger appearance, size, or location.

\vspace{1mm}
\noindent{\bf Defenses Against Backdoor Attacks.}
Existing defenses can be broadly categorized according to the stage of the
model lifecycle at which they operate.

\noindent\underline{\textbf{Pre-training}} defenses aim to identify and remove
poisoned samples before model training. Representative approaches exploit
activation clustering~\cite{chen2018detecting}, feature
consistency~\cite{hayase2021spectre}, or visual--linguistic
misalignment~\cite{zhu2023vdc} to distinguish suspicious samples from benign
training data. Although effective when the training set is available, these
methods are difficult to apply when users receive already trained and
potentially compromised models.

\noindent\underline{\textbf{Training-time}} defenses attempt to prevent
backdoor learning during model optimization. Anti-Backdoor Learning
(ABL)~\cite{li2021anti} and Decouple-based Backdoor Defense
(DBD)~\cite{gao2023backdoor}, for example, identify suspicious training
samples and modify the optimization process to suppress their influence.
Other approaches introduce controlled poisoning or defensive perturbations
to improve resistance against backdoor injection~\cite{chen2022effective}.
These methods, however, require access to and control over the training
procedure and therefore cannot directly protect already deployed
query-only models.

\noindent\underline{\textbf{Post-training}} defenses analyze or sanitize a
trained model before deployment. Neural Attention
Distillation~\cite{li2021neural} and ScaleUp~\cite{guo2023scale}, for
instance, exploit attention patterns or prediction consistency to identify
and mitigate backdoor behavior. While these approaches avoid retraining from
scratch, they may still depend on auxiliary data, model access, or trusted
reference information that is unavailable when the deployed model is
accessible only through its prediction interface.

\noindent\underline{\textbf{Inference-time}} defenses operate directly on
incoming samples after deployment, making them particularly relevant to
practical black-box settings. Detection-oriented methods such as
DBDR~\cite{shao2021bddr}, Beatrix~\cite{ma2022beatrix},
TeCo~\cite{liu2023detecting}, and FLARE~\cite{hou2025flare} identify
suspicious inputs or backdoor behavior, while restoration-oriented methods
such as Orion~\cite{huang2023orion}, ZIP~\cite{shi2023black}, and
SampDetox~\cite{yang2024sampdetox} recover benign predictions through input
transformation or feature purification. Recent work by Xue \emph{et
al.}~\cite{xue2026unified} introduces \emph{Backdoor Trigger Segmentation
(BTS)} to localize diverse trigger regions, highlighting the importance of
region-level analysis. However, existing approaches typically detect
suspicious inputs, localize triggers, or transform broad portions of an
input, without jointly verifying which regions causally influence the
backdoored prediction and selectively purifying them. In contrast,
\textit{TRIM} uses localization as an intermediate step toward prediction
restoration: it filters candidate regions, selectively reconstructs them,
and verifies prediction changes through black-box queries before purifying
only influential regions. This targeted intervention minimizes unnecessary
modification of benign content while supporting diverse trigger patterns
without requiring model internals, training data, or clean validation
samples.

\section{Problem Statement, Threat Model, and Defense Assumptions}
\label{sec:problmDefinition}

\noindent\textbf{Problem Statement.}
In supervised learning, a DNN $f_\Theta$ is trained on
$D=\{(x_i,y_i)\}_{i=1}^{N}$ to map an input $x$ to its ground-truth label
$y\in Y$. In a backdoor attack, an adversary injects a trigger $b$ into a
subset of training samples, yielding compromised parameters $\Theta_b$.
For a triggered input $x_b=x+b$, the backdoored model satisfies
\[
f_{\Theta_b}(x_b)=y', \qquad y'\neq y,
\]
while maintaining high accuracy on benign inputs. The attack therefore
associates $b$ with an attacker-specified target while concealing the
backdoor under normal operation.

\vspace{1mm}
\noindent\textbf{Threat Model.} We consider a practical poisoning threat model~\cite{gu2017badnets,li2022backdoor} in which an adversary poisons a subset $D_b\subset D$ of the training data, with poisoning ratio $r=|D_b|/|D|$. The resulting model $f_{\Theta_b}$ behaves normally on clean inputs but exhibits attacker-controlled behavior on triggered inputs. Poisoning occurs before deployment, reflecting scenarios involving compromised datasets or untrusted pre-trained models, with no attacker interaction required at inference.

\vspace{1mm}
\noindent\textbf{Backdoor Defense Assumptions.}
Inference-time defenses can be categorized according to defender access:
\begin{itemize}[leftmargin=*]
    \item \textbf{White-box:} Full access to model parameters $\Theta$ and
    potentially training data $D$, enabling
    $g(x,\Theta,D)$ using internal information.

    \item \textbf{Gray-box:} Limited access, such as partial parameters
    $\Theta_{\mathrm{partial}}$, validation data $D_{\mathrm{val}}$, or a
    surrogate model, yielding
    $g(x,\Theta_{\mathrm{partial}},D_{\mathrm{val}})$.

    \item \textbf{Black-box:} No access to model internals or training data;
    the defender interacts only through input--output queries
    $\hat{y}=f(x)$, and the defense relies on observable model behavior,
    i.e., $g(x,f(x))$.
\end{itemize}

\noindent
\textit{TRIM} operates in the strict black-box setting: it detects and
removes triggers at inference time using only model queries, without access
to training data, model internals, clean validation samples, or prior
knowledge of the trigger.


\vspace{-0.25cm}
\section{Methodology}\label{sec:method}

\begin{figure}[!t]
    \centering
    \includegraphics[width=\textwidth]{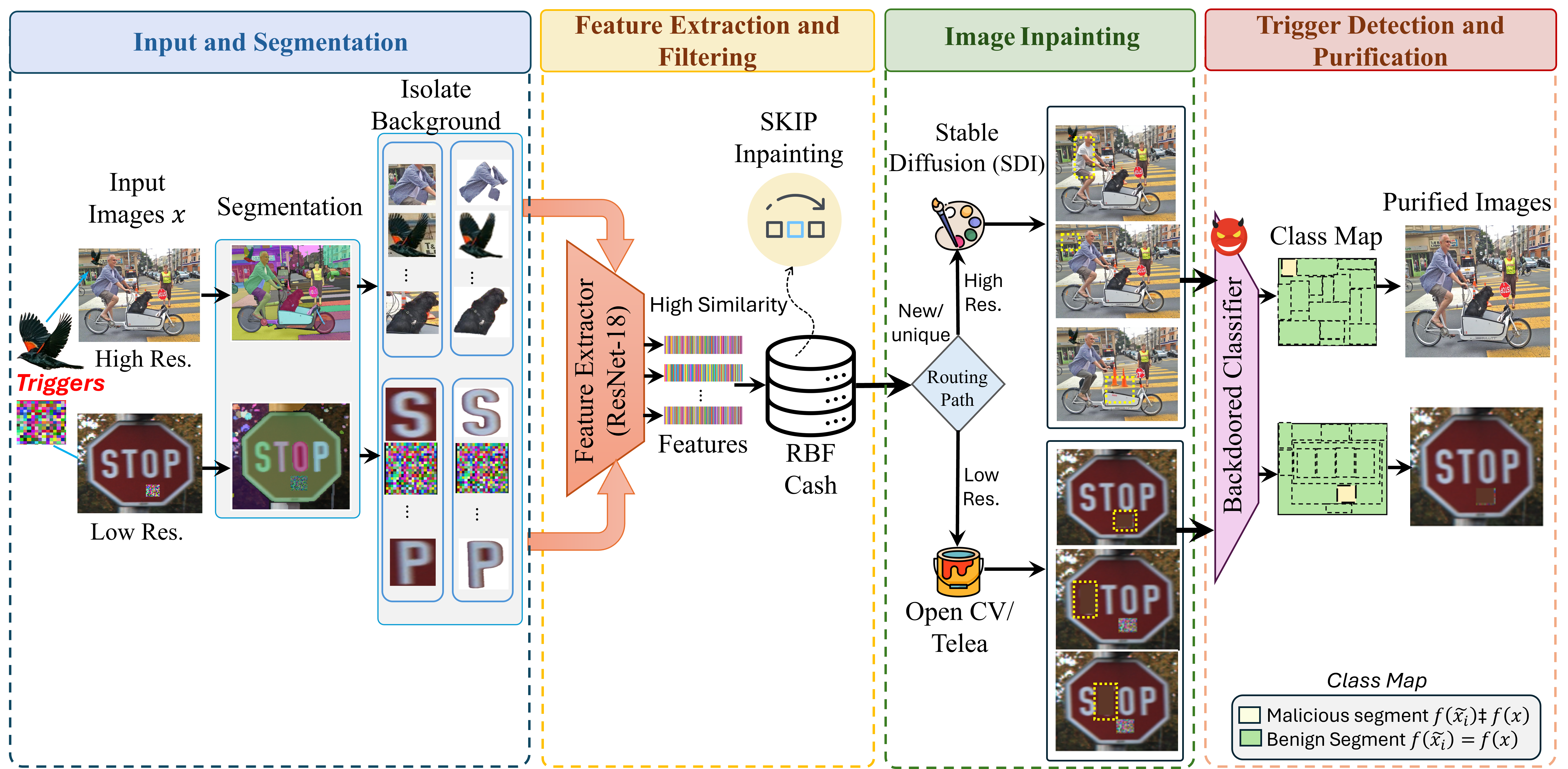}
    \vspace{-4mm}
    \caption{Overview of \textit{TRIM}. The input is segmented into regions
    and encoded into feature representations. Candidate regions are filtered,
    selectively reconstructed, and verified using black-box prediction
    changes. Verified trigger regions are jointly purified, while region
    features are cached to accelerate subsequent inference.}
    \label{fig:Methodology}
    \vspace{-4mm}
\end{figure}

We propose \textit{TRIM}, a model-agnostic, inference-time defense for
detecting and removing backdoor triggers from compromised DNNs. \textit{TRIM}
operates entirely in a \emph{black-box} setting, requiring only model
predictions without access to architecture, parameters, training data, clean
validation samples, or trigger characteristics. Rather than transforming the
entire input, \textit{TRIM} identifies candidate regions, verifies their
influence on the model prediction, and selectively purifies trigger-related
content. As shown in~\Cref{fig:Methodology}, the pipeline comprises three
stages:

\begin{enumerate}[label={\bf \arabic*.}]

    \item \textbf{\em Region Segmentation and Feature Extraction:}
    The input image $x$ is decomposed into semantic regions, and a pre-trained
    encoder (e.g., ResNet-18) extracts a compact feature representation for
    each region. These embeddings support candidate filtering and comparison
    with previously analyzed benign and trigger-related regions.

    \item \textbf{\em Candidate Reconstruction and Verification:}
    Large semantic regions likely corresponding to primary objects or
    background content are excluded, while feature-based filtering avoids
    redundant evaluation of previously analyzed patterns. Each remaining
    candidate is selectively suppressed and reconstructed to produce an
    intervention image. Lightweight inpainting is used for low-resolution
    inputs, while generative inpainting (e.g., Stable Diffusion) supports
    context-preserving reconstruction of high-resolution inputs. Each variant
    is then queried against the black-box classifier to measure the region's
    influence on its prediction.

    \item \textbf{\em Trigger Detection and Selective Purification:}
    The prediction of each intervention image is compared with that of the
    original input. A prediction change after candidate filtering indicates
    that the suppressed region influences the model decision and is therefore flagged as suspicious and potentially trigger-related. Verified masks are merged into a single
    purification mask, and the original image is purified once over these
    regions. Features of analyzed benign and trigger-related regions are
    cached to reduce redundant processing in subsequent inputs.

\end{enumerate}

These stages enable localized trigger detection and purification using only
black-box feedback. By separating candidate filtering, region verification,
and final purification, \textit{TRIM} limits modification of benign content
while supporting diverse trigger configurations.

 



\vspace{-0.1cm}
\subsection{Region-Based Feature Extraction}
\label{sec:region_feature}

Given an input image $x$, \textit{TRIM} constructs region-level
representations through three steps (see~\Cref{fig:Methodology}): segmentation,
background isolation, and feature extraction.

\noindent{\bf 4.1.1 Semantic Segmentation.}
We decompose $x$ into candidate regions using the Segment Anything Model
(SAM)~\cite{kirillov2023segment}. SAM's \textit{SamAutomaticMaskGenerator}
operates in prompt-free mode, providing a trigger-agnostic partition without
prior knowledge of trigger appearance or location. Given $N$ regularly
sampled prompt points $P_r=\{p_r^i\}_{i=1}^{N}$, SAM produces candidate masks
$\mathcal{M}_i=\{\hat{M}_{i,k}\}_{k=1}^{K_i}$ for each point. The complete
mask set is
\begin{equation}
    \mathcal{M}
    =
    \bigcup_{i=1}^{N}\mathcal{M}_i
    =
    \{\hat{M}_{i,k}\mid i=1,\ldots,N;\;k=1,\ldots,K_i\}.
\end{equation}
We remove very small or largely contained masks and consolidate highly
overlapping regions, yielding a compact candidate set
$\mathcal{S}=\{s_j\}_{j=1}^{J}$ for subsequent analysis.

\vspace{1mm}
\noindent{\bf 4.1.2 Background Isolation.}
For each segment $s_j$ with binary mask $M_j$, we suppress surrounding
content before feature extraction:
\begin{equation}
    x_j^{\mathrm{seg}} = x \odot M_j,
\end{equation}
where $\odot$ denotes element-wise multiplication. This reduces background
interference and emphasizes region-specific characteristics. Candidate
segments need not correspond to predefined semantic classes and may represent
objects, object parts, textures, or anomalous patterns.

\vspace{1mm}
\noindent{\bf 4.1.3 Segment-Level Feature Extraction.}
Each isolated segment is encoded using a pre-trained feature encoder
$g(\cdot)$ (e.g., ResNet-18):
\begin{equation}
    z_j =
    \frac{g(x_j^{\mathrm{seg}})}
    {\|g(x_j^{\mathrm{seg}})\|_2+\epsilon},
\end{equation}
where features are extracted before the final classification layer and
$\epsilon$ ensures numerical stability. The normalized embeddings support
candidate filtering and comparison with cached benign and trigger-related
regions. Importantly, $g(\cdot)$ is independent of the protected classifier,
preserving \textit{TRIM}'s black-box assumption without requiring access to
the victim model's architecture, parameters, or internal activations.

\subsection{Segment-Wise Image Inpainting}
\label{sec:inpainting}

Given the candidate regions from the previous stage, \textit{TRIM} generates
intervention images by selectively reconstructing one region at a time. For
each segment $s_j$ with mask $M_j$, an inpainted variant $\hat{x}_j$ replaces
only the masked region while preserving the remaining content. These variants
serve as counterfactual interventions for evaluating whether suppressing a
region changes the black-box model's prediction, rather than as final purified
inputs. The reconstruction method is selected according to image resolution
to balance visual fidelity and efficiency.

\vspace{1mm}
\noindent{\bf High-Resolution Inputs.}
For high-resolution images, we use Stable Diffusion Inpainting
(SDI)~\cite{rombach2022high} to generate contextually plausible replacements.
Given image $x$, mask $M_j\in\{0,1\}^{H\times W}$, and a generic prompt $P$,
the intervention is
\begin{equation}
    \hat{x}_j=\mathcal{D}_{\theta}(x,M_j,P),
\end{equation}
where $\mathcal{D}_{\theta}$ denotes the diffusion inpainting model. Pixels
outside $M_j$ remain unchanged, while masked content is replaced:
\begin{equation}
    \hat{x}_j(p)=
    \begin{cases}
        x(p), & M_j(p)=0,\\
        \mathcal{D}_{\theta}(x,M_j,P)(p), & M_j(p)=1.
    \end{cases}
\end{equation}
Rather than recovering the original region, which may preserve the trigger,
SDI generates a plausible replacement that suppresses its influence. This is
well suited to high-resolution natural images such as
Caltech-101~\cite{fei2004learning} and ImageNet~\cite{deng2009imagenet}.

\vspace{1mm}
\noindent{\bf Low-Resolution Inputs.}
For low-resolution images, diffusion reconstruction can introduce unnecessary
overhead with limited contextual benefit. We therefore use the lightweight
Telea algorithm~\cite{telea2004image}, implemented by
\texttt{cv2.inpaint}:
\begin{equation}
    \hat{x}_j=
    \operatorname{Inpaint}_{\mathrm{Telea}}(x,M_j,r),
\end{equation}
where $r$ is the inpainting radius. Masked pixels are reconstructed from
neighboring known pixels $\mathcal{N}_r(p)$:
\begin{equation}
    \hat{x}_j(p)=
    \begin{cases}
        x(p), & M_j(p)=0,\\[1mm]
        \displaystyle
        \frac{\sum_{q\in\mathcal{N}_r(p)}w(p,q)x(q)}
             {\sum_{q\in\mathcal{N}_r(p)}w(p,q)},
        & M_j(p)=1,
    \end{cases}
\end{equation}
where $w(p,q)$ denotes the spatial weights used to propagate neighboring
information. This provides an efficient intervention for low-resolution
datasets such as CIFAR-10~\cite{krizhevsky2009learning}.

\vspace{1mm}
For $\mathcal{S}=\{s_j\}_{j=1}^{J}$, the procedure produces
$\hat{\mathcal{X}}=\{\hat{x}_j\}_{j=1}^{J}$. \textit{TRIM} subsequently
queries the black-box classifier with these interventions and uses the
prediction responses to verify trigger-related regions before final
purification.

\subsection{Trigger Detection, Isolation, and Purification}
\label{sec:trigger_detection}

This stage identifies candidate regions associated with backdoor behavior and
selectively removes their influence from the input. \textit{TRIM} performs
three operations: (i) candidate filtering and region-level verification,
(ii) feature-based caching to avoid redundant analysis, and (iii) mask
aggregation followed by selective purification.

\vspace{1mm}
\noindent{\bf 1. Candidate Filtering and Trigger Verification.}
A prediction change after suppressing an image region does not necessarily
indicate a backdoor trigger, since removing task-relevant content can also
alter the classifier's decision. \textit{TRIM} therefore filters candidate
regions before prediction-based verification. Specifically, large regions
likely to correspond to primary objects or dominant background content are
excluded from trigger analysis. When cached representations are available,
candidate embeddings are additionally compared with previously analyzed
benign and trigger-related regions to avoid redundant evaluation.

For each remaining candidate segment $s_i$, we use the intervention image
$\tilde{x}_i$ generated in Sec.~\ref{sec:inpainting}, in which only $s_i$ is
suppressed and reconstructed. Let $f(\cdot)$ denote the protected black-box
classifier. We obtain
\begin{equation}
    y_0=f(x), \qquad y_i=f(\tilde{x}_i),
\end{equation}
where $y_0$ and $y_i$ are the predictions for the original and intervened
inputs, respectively. We define the region-level verification indicator as
\begin{equation}
    d_i=\mathbf{1}\!\left[y_i\neq y_0\right],
\end{equation}
where $\mathbf{1}[\cdot]$ denotes the indicator function. A candidate with
$d_i=1$ is flagged as suspicious because suppressing the corresponding region
changes the model decision, providing evidence that the region influences the
backdoored prediction. Candidates with $d_i=0$ are not flagged by the current
intervention. Importantly, prediction change is used as a \emph{verification
signal after candidate filtering}, rather than as the sole criterion for
trigger localization.

\vspace{1mm}
\noindent{\bf 2. Trigger Isolation and Feature Caching.}
To reduce redundant computation across subsequent inputs, \textit{TRIM}
maintains two feature-memory pools, $\mathcal{B}$ and $\mathcal{T}$, containing
embeddings of previously analyzed benign and trigger-related regions,
respectively. Given a candidate embedding $z_i$ and a cached representation
$z_j$, their similarity is measured using the Radial Basis Function (RBF)
kernel:
\begin{equation}
    \kappa(z_i,z_j)
    =
    \exp\!\left(-\gamma\|z_i-z_j\|_2^2\right),
\end{equation}
where $\gamma$ controls sensitivity to feature distance. For a memory pool
$\mathcal{C}\in\{\mathcal{B},\mathcal{T}\}$, we compute
\begin{equation}
    \rho_i(\mathcal{C})
    =
    \max_{z_j\in\mathcal{C}}\kappa(z_i,z_j).
\end{equation}
If the similarity to a cached representation exceeds a threshold $\tau$
($\tau=0.9$ in our experiments), its previous assignment can be reused,
avoiding redundant reconstruction and classifier queries. Otherwise, the
candidate undergoes region-level verification, and its embedding is
subsequently added to the corresponding memory pool. The cache therefore
becomes progressively more informative as additional regions are analyzed,
reducing repeated processing of recurring visual patterns.

\vspace{1mm}
\noindent{\bf 3. Mask Aggregation and Selective Purification.}
After region-level verification, the masks of all flagged regions are combined
into a single purification mask:
\begin{equation}
    M_{\mathrm{pur}}
    =
    \bigvee_{i:\,d_i=1} M_i,
\end{equation}
where $\bigvee$ denotes the pixel-wise union of the verified masks. The
intermediate variants $\tilde{x}_i$ are used only to assess individual region
influence; they are not retained as final purified inputs. Instead,
\textit{TRIM} performs a \emph{single final purification} of the original
image over $M_{\mathrm{pur}}$, thereby limiting unnecessary modification of
benign content.

The purification operator is selected according to the input resolution.
Stable Diffusion Inpainting (SDI) is used for high-resolution natural images,
whereas \texttt{cv2.inpaint} provides a lightweight alternative for
low-resolution inputs. The final purified image is
\begin{equation}
    x_{\mathrm{pur}}
    =
    \mathcal{P}(x,M_{\mathrm{pur}}),
\end{equation}
where $\mathcal{P}(\cdot)$ denotes the selected reconstruction operator.
The protected classifier then produces the final prediction
$f(x_{\mathrm{pur}})$. During evaluation, purification is considered
successful when this prediction recovers the ground-truth class.

To limit over-sanitization, regions exceeding a predefined area threshold are
excluded from verification and purification because they are more likely to
represent primary objects or dominant scene content. We select this threshold
empirically through the sensitivity analysis reported in
Table~\ref{tab:segmentation}.

\vspace{1mm}
\noindent{\bf Multi-Trigger Behavior and Limitation.}
When multiple triggers occur in the same input, \textit{TRIM} evaluates their
regions independently. If one trigger dominates the model's prediction,
suppressing a weaker trigger alone may not cause a prediction change, leaving
it undetected in that image. However, once the dominant trigger is detected,
its representation is cached; in subsequent images containing the same
dominant pattern, \textit{TRIM} can reuse the cached decision and focus
verification on the remaining candidates, increasing the likelihood of
detecting weaker or non-dominant triggers. For attacks whose trigger consists
of multiple spatially distributed components, removing even a subset of these
components can weaken the collective trigger effect sufficiently to alter the
prediction, allowing the corresponding regions to be identified and purified.
Nevertheless, interacting non-dominant triggers that individually produce no
measurable prediction change within a single previously unseen input remain a
limitation of the current region-wise verification strategy.

\vspace{1mm}
\noindent{\bf Deployment Efficiency.}
Although \textit{TRIM} involves segmentation, region-level reconstruction,
and black-box classifier queries, candidate filtering and feature caching
reduce redundant computation, while final purification is restricted to the
union of verified regions. The framework also supports different
computational operating points: high-quality segmentation and diffusion-based
reconstruction prioritize reconstruction fidelity, whereas lightweight
components such as MobileSAM~\cite{zhang2023faster} and
\texttt{cv2.inpaint} provide substantially faster alternatives. We therefore
characterize \textit{TRIM} as \emph{deployment-tunable} rather than
universally real-time. Detailed latency and computational analyses are
reported in Table \ref{tab:TRIM_cost_breakdown}.

 \vspace{-3mm}

\section{Experiments and Results}
\label{sec:exp}


We conduct extensive experiments to address five questions:
\textbf{(Q1)} How effectively does \textit{TRIM} suppress diverse backdoor attacks while preserving clean accuracy?
\textbf{(Q2)} Does \textit{TRIM} generalize across attack objectives, trigger sizes, locations, and generation mechanisms?
\textbf{(Q3)} How reliably does its region-level intervention identify and purify trigger-related regions while preserving benign content?
\textbf{(Q4)} How do the segmentation threshold and feature caching affect robustness and efficiency?
\textbf{(Q5)} What is the computational overhead of the complete defense pipeline?

\vspace{-3mm}
\subsection{Experimental Settings}

\noindent\textbf{Datasets.}
We evaluate \textit{TRIM} on CIFAR-10~\cite{krizhevsky2009learning} and
ImageNet-10~\cite{imagenette2020}, covering low- and high-resolution settings.
CIFAR-10 images are bilinearly upsampled from $32\times32$ to $224\times224$
to provide sufficient granularity for segmentation. ImageNet-10 contains
13,394 images (9,469 training and 3,925 validation) resized to
$256\times256$. Together, these datasets evaluate \textit{TRIM} across
different resolutions and visual complexities.

\vspace{0.5mm}
\noindent\textbf{Backdoor Attacks.}
We consider diverse trigger constructions: \textit{BadNets}~\cite{gu2017badnets},
\textit{Blended}~\cite{chen2017targeted}, \textit{Label-Consistent
(LC)}~\cite{turner2019label}, \textit{Physical}~\cite{wenger2021backdoor},
\textit{Dynamic}~\cite{salem2022dynamic}, \textit{Distributed (DBA)}
~\cite{xie2020dba}, \textit{Input-Aware}~\cite{nguyen2020input}, and
\textit{VSSC}~\cite{wang2023robust}, covering localized, blended, physical,
dynamic, distributed, and sample-specific triggers. We also evaluate a
\textit{multi-trigger} setting containing two spatially separated triggers to
assess the region-wise verification limitation discussed in
Sec.~\ref{sec:trigger_detection}.

\vspace{0.5mm}
\noindent\textbf{Attack Objectives.}
We evaluate All-to-One (A2O), All-to-All (A2A), and untargeted (UT) attacks.
A2O maps triggered samples to one target class, A2A maps each source class to
a predefined incorrect class, and UT considers any incorrect prediction
successful. These settings evaluate robustness across different attack
objectives.

\vspace{0.5mm}
\noindent\textbf{Backdoored Model Preparation.}
Unless otherwise specified, $10\%$ of training samples are poisoned with the
corresponding trigger and jointly trained with clean samples. ResNet-50 is the
primary protected classifier. During inference, \textit{TRIM} accesses only
its predictions, without parameters, gradients, activations, or training data.

\vspace{0.5mm}
\noindent\textbf{Evaluation Metrics.}
We primarily report \textit{Attack Success Rate (ASR)} and \textit{Clean
Accuracy (CA)}. ASR measures successful attacks on triggered samples, while CA
measures accuracy on clean inputs. We additionally report \textit{False
Positive Rate (FPR)}, \textit{Detection Accuracy (DA)}, \textit{Purified
Accuracy (PA)}, and \textit{Artifact Ratio (AR)}. AR measures visual artifacts
introduced on benign inputs:
\begin{equation}
    \mathrm{AR}
    =\frac{N_{\mathrm{artifact}}}{N_{\mathrm{clean}}}\times100\%.
\end{equation}

\vspace{0.5mm}
\noindent\textbf{Implementation Details.}
We use SAM ViT-H for segmentation and pretrained ResNet-18 for independent
region-level feature encoding. Embeddings are $\ell_2$-normalized before RBF
comparison. Unless otherwise specified, the cache threshold is $\tau=0.9$
and the segment-area threshold is $0.25$ (Table~\ref{tab:segmentation}).
Low-resolution images use \texttt{cv2.inpaint}, whereas high-resolution
images use Stable Diffusion inpainting.

\vspace{0.5mm}
\noindent\textbf{Environmental Setup.}
Experiments use Ubuntu~20.04, an NVIDIA RTX~4090 (24\,GB), Intel i9-13900K
(24 cores), and 128\,GB RAM, with Python~3.10, PyTorch~2.1.0/CUDA~12.1,
OpenCV~4.8.0, and SAM ViT-H.

\vspace{-2mm}

\subsection{Comparison with Baseline Defenses}
\label{sec:baseline_comparison}

Table~\ref{tab:defense_comparison} compares \textit{TRIM} with ShrinkPad, ZIP, and FLARE under the standard attack configuration. \textit{TRIM} consistently achieves a strong robustness--utility trade-off: on CIFAR-10, it reduces average ASR from $97.94\%$ to $3.77\%$ while retaining $85.83\%$ CA, and on ImageNet-10, from $78.45\%$ to $5.06\%$ while maintaining $76.81\%$ CA. In contrast, competing defenses can substantially sacrifice clean accuracy; for example, under ImageNet-10 Physical, FLARE achieves $6.37\%$ ASR with only $16.33\%$ CA, whereas \textit{TRIM} achieves $5.18\%$ ASR with $78.37\%$ CA. Although ZIP performs better for CIFAR-10 Blended ($3.62\%$ vs.\ $6.13\%$ ASR), \textit{TRIM} provides more consistent performance across attacks. To assess statistical robustness, we repeat all \textit{TRIM} experiments over five seeds and report mean$\pm$standard deviation. The low variability, e.g., $87.22\pm0.58\%$ CA/$3.30\pm0.41\%$ ASR on CIFAR-10 BadNets and $78.68\pm0.82\%$/$5.98\pm0.61\%$ on ImageNet-10 BadNets, further indicates stable performance across runs.
\begin{table*}[t]
\centering
\caption{Comparison of TRIM vs. baselines on various attack modes for CIFAR-10 and ImageNet-10 with $3{\times}3$-pixel triggers. Bold indicates best performance ($\boldsymbol{\uparrow}$ higher CA, $\boldsymbol{\downarrow}$ lower ASR). All values are percentages (\%). TRIM results are reported as
mean$\pm$standard deviation over five independent random seeds.}

\label{tab:defense_comparison}

\setlength{\tabcolsep}{4.2pt}
\renewcommand{\arraystretch}{1.10}

\resizebox{\textwidth}{!}{%
\begin{tabular}{ll|cc|cc|cc|cc|cc}
\toprule

\multirow{2}{*}{\textbf{Dataset}}
& \multirow{2}{*}{\textbf{Attack}}
& \multicolumn{2}{c|}{\textbf{No Defense}}
& \multicolumn{2}{c|}{\textbf{ShrinkPad}}
& \multicolumn{2}{c|}{\textbf{ZIP}}
& \multicolumn{2}{c|}{\textbf{FLARE}}
& \multicolumn{2}{c}{\textbf{TRIM (Ours)}} \\

\cmidrule(lr){3-4}
\cmidrule(lr){5-6}
\cmidrule(lr){7-8}
\cmidrule(lr){9-10}
\cmidrule(lr){11-12}

&
& \textbf{CA}$\uparrow$ & \textbf{ASR}$\downarrow$
& \textbf{CA}$\uparrow$ & \textbf{ASR}$\downarrow$
& \textbf{CA}$\uparrow$ & \textbf{ASR}$\downarrow$
& \textbf{CA}$\uparrow$ & \textbf{ASR}$\downarrow$
& \textbf{CA}$\uparrow$ & \textbf{ASR}$\downarrow$ \\

\midrule


\multirow{6}{*}{\rotatebox{90}{\textbf{CIFAR-10}}}

& BadNets
& 91.29 & 97.17
& 70.37 & 7.42
& 78.84 & 5.56
& 82.31 & 12.09
& $\mathbf{87.22\pm0.58}$
& $\mathbf{3.30\pm0.41}$ \\

& Blended
& 90.98 & 100.00
& 41.66 & 7.10
& 82.54 & \textbf{3.62}
& 72.17 & 17.80
& $\mathbf{87.87\pm0.64}$
& $6.13\pm0.52$ \\

& Physical
& 93.72 & 99.99
& \textbf{86.12} & 10.13
& 77.32 & 23.66
& 73.34 & 17.89
& $84.50\pm0.71$
& $\mathbf{4.70\pm0.47}$ \\

& LC
& 92.03 & 99.01
& \textbf{85.11} & 9.13
& 80.13 & 88.20
& 57.14 & 11.30
& $82.53\pm0.66$
& $\mathbf{1.16\pm0.28}$ \\

& Dynamic
& 91.41 & 93.53
& 83.42 & 10.34
& 82.11 & 55.13
& 67.13 & 34.00
& $\mathbf{87.01\pm0.73}$
& $\mathbf{3.56\pm0.44}$ \\

\cmidrule(lr){2-12}

& \textbf{Average}
& 91.89
& \cellcolor{red!10}{\textbf{97.94}}
& 73.34 & 8.82
& 80.19 & 35.23
& 70.42 & 18.62
& \cellcolor{green!20}{\textbf{$\mathbf{85.83\pm0.66}$}}
& \cellcolor{green!20}{\textbf{$\mathbf{3.77\pm0.42}$}} \\

\midrule


\multirow{6}{*}{\rotatebox{90}{\textbf{ImageNet-10}}}

& BadNets
& 85.12 & 96.03
& 72.15 & 8.43
& 67.67 & 13.23
& \textbf{81.20} & 19.45
& $78.68\pm0.82$
& $\mathbf{5.98\pm0.61}$ \\

& Blended
& 83.26 & 99.97
& 65.02 & 19.50
& 63.15 & 88.12
& 68.14 & 20.03
& $\mathbf{77.12\pm0.91}$
& $\mathbf{6.12\pm0.67}$ \\

& Physical
& 79.75 & 88.64
& 77.09 & 76.15
& 61.19 & 20.03
& 16.33 & 6.37
& $\mathbf{78.37\pm0.88}$
& $\mathbf{5.18\pm0.56}$ \\

& LC
& 75.16 & 97.16
& 37.89 & 77.78
& 54.16 & 81.36
& 53.11 & 9.55
& $\mathbf{71.50\pm0.95}$
& $\mathbf{2.10\pm0.36}$ \\

& Dynamic
& 84.63 & 10.46
& 62.80 & 10.11
& 61.28 & 10.31
& \textbf{81.74} & 12.20
& $78.37\pm1.02$
& $\mathbf{5.90\pm0.72}$ \\

\cmidrule(lr){2-12}

& \textbf{Average}
& 81.58
& \cellcolor{red!10}{\textbf{78.45}}
& 62.99 & 38.39
& 61.49 & 42.61
& 60.20 & 13.52
& \cellcolor{green!20}{\textbf{$\mathbf{76.81\pm0.92}$}}
& \cellcolor{green!20}{\textbf{$\mathbf{5.06\pm0.58}$}} \\

\bottomrule
\end{tabular}%
}

\vspace{-3mm}
\end{table*}

\subsection{Robustness Across Attack Objectives}
\label{sec:attack_modes}

\noindent\textbf{CIFAR-10.}
Table~\ref{tab:defense_comparison_cifar10_modes} evaluates whether the
effectiveness of \textit{TRIM} depends on the attacker's target-label
strategy. Across A2O, A2A, and UT attacks, \textit{TRIM} achieves average
ASRs of $4.71\%$, $4.56\%$, and $4.04\%$, respectively, while maintaining
average CAs of $86.53\%$, $84.86\%$, and $80.03\%$. Thus, its effectiveness
does not rely on a fixed A2O target mapping.

Several individual cases further illustrate the robustness--utility
trade-off. ShrinkPad attains higher CA for Physical A2O and Physical UT, but
its corresponding ASRs increase to $10.13\%$ and $27.50\%$, respectively.
ZIP obtains the lowest ASR for Blended A2O ($3.62\%$) and Blended A2A
($4.13\%$), but its robustness varies considerably across the remaining
settings. In contrast, \textit{TRIM} maintains ASR below $8.1\%$ in every
reported CIFAR-10 attack-mode configuration.

\noindent\textbf{ImageNet-10.}
Table~\ref{tab:imagenet_defense_comparison} shows a similar trend at higher
resolution. \textit{TRIM} achieves average ASRs of $5.76\%$ and $6.90\%$
under A2O and A2A, respectively, while retaining average CAs of $78.06\%$
and $72.29\%$. In comparison, FLARE attains average CAs of only $55.22\%$
and $55.64\%$ in the corresponding settings. These results suggest that
region-selective intervention remains effective across different attack
objectives and image resolutions.
\vspace{-3mm}

\begin{table*}[!t]
\centering
\caption{Comparison of TRIM vs. baselines on various attack modes for CIFAR-10. Bold indicates best performance ($\boldsymbol{\uparrow}$ higher CA, $\boldsymbol{\downarrow}$ lower ASR). All values are percentages (\%).}

\resizebox{\linewidth}{!}{ 
\begin{tabular}{l|l|cc|cc|cc|cc|cc}
\toprule
\multirow{2.5}{*}{\rotatebox{90}{\textbf{Mode}}} & \multirow{2}{*}{\textbf{Attack}} 
& \multicolumn{2}{c|}{\textbf{No Defense}} 
& \multicolumn{2}{c|}{\textbf{ShrinkPad}} 
& \multicolumn{2}{c|}{\textbf{ZIP}} 
& \multicolumn{2}{c|}{\textbf{FLARE}} 
& \multicolumn{2}{c}{\textbf{TRIM (Ours)}} \\
\cmidrule(lr){3-4} \cmidrule(lr){5-6} \cmidrule(lr){7-8} \cmidrule(lr){9-10} \cmidrule(lr){11-12}
& & \textbf{CA} $\boldsymbol{\uparrow}$ & \textbf{ASR} $\boldsymbol{\downarrow}$ 
  & \textbf{CA}$\boldsymbol{\uparrow}$ & \textbf{ASR} $\boldsymbol{\downarrow}$ 
  & \textbf{CA}$\boldsymbol{\uparrow}$ & \textbf{ASR} $\boldsymbol{\downarrow}$ 
  & \textbf{CA}$\boldsymbol{\uparrow}$ & \textbf{ASR} $\boldsymbol{\downarrow}$ 
  & \textbf{CA}$\boldsymbol{\uparrow}$ & \textbf{ASR} $\boldsymbol{\downarrow}$\\
\midrule

\multirow{4}{*}{\rotatebox{90} {\bf A2O}}
& \text{BadNets} & 91.29 & 97.17 & 70.37 & 7.42  & 78.84 & 5.56  & 82.31 & 12.09 & \textbf{87.22} & \textbf{3.30} \\
& \text{Blended}    & 90.98 & 100.00 & 41.66 & 7.10  & 82.54 & \textbf{3.62}  & 72.17 & 17.80 & \textbf{87.87} & 6.13 \\
& \text{Physical}   & 93.72 & 99.99 & \textbf{86.12} & 10.13 & 77.32 & 23.66 & 73.34 & 17.89 & 84.50 & \textbf{4.70} \\
\cmidrule(lr){2-12}
& \textit{Average} & \cellcolor{red!10}{\bf91.99} & 99.05 & 66.05 & 8.88 & 79.57 & 10.95 & 75.94 & 15.93 & \cellcolor{green!20}{\bf 86.53} & \cellcolor{green!20}{\bf4.71} \\

\hline\hline

\multirow{4}{*}{\rotatebox{90} {\bf A2A}}
& \text{BadNets}    & 90.88 & 89.15 & 70.10 & 8.60  & 80.76 & 15.24 & 81.95 & 13.07 & \textbf{87.00} & \textbf{3.41} \\
& \text{Blended}    & 90.24 & 83.67 & 49.93 & 40.10 & 77.12 & \textbf{4.13}  & 72.17 & 17.80 & \textbf{81.30} & 8.07 \\
& \text{Physical}   & 92.11 & 91.40 & 84.98 & 18.60 & 79.17 & 33.56 & 74.00 & 13.89 & \textbf{86.29} & \textbf{2.20} \\
\cmidrule(lr){2-12}
& \textit{Average} & \cellcolor{red!10}{\bf91.08} & 88.07 & 68.34 & 22.43 & 79.68 & 18.98 & 76.04 & 14.92 & \cellcolor{green!20}{\bf84.86} & \cellcolor{green!20}{\bf4.56} \\

\hline\hline

\multirow{4}{*}{\rotatebox{90} {\bf UT}}
& \text{BadNets}    & 89.45 & 89.31 & 69.54 & 13.60 & \textbf{79.55} & 17.93 & 81.56 & 12.69 & 76.15 & \textbf{2.32} \\
& \text{Blended}    & 90.12 & 80.13 & 44.17 & 47.10 & 78.32 & 9.88  & 74.11 & 14.74 & \textbf{85.76} & \textbf{6.93} \\
& \text{Physical}   & 89.20 & 93.45 & \textbf{81.80} & 27.50 & 79.31 & 30.19 & 74.09 & 12.32 & \textbf{78.19} & \textbf{2.87} \\
\cmidrule(lr){2-12}
& \textit{Average} & \cellcolor{red!10}{\bf89.59} & 87.63 & 66.23 & 29.40 & 79.06 & 19.33 & 76.59 & 13.25 & \cellcolor{green!20}{\bf80.03} & \cellcolor{green!20}{\bf4.04} \\

\bottomrule
\end{tabular}}
\label{tab:defense_comparison_cifar10_modes}\vspace{-1mm}
\end{table*}

\begin{table}[t!]
\centering
\scriptsize
\renewcommand{\arraystretch}{1.15}
\setlength{\tabcolsep}{2.5pt}

\caption{Comparison of TRIM vs. baselines on various attack modes for ImageNet-10. Bold indicates best performance ($\uparrow$ higher CA, $\downarrow$ lower ASR). All values are percentages (\%).}

\resizebox{\columnwidth}{!}{
\begin{tabular}{l|c||cc||cc||cc||cc||cc}
\hline
\multirow{2}{*}{Mode} & \multirow{2}{*}{Attack} 
& \multicolumn{2}{c||}{No Def.} 
& \multicolumn{2}{c||}{ShrinkPad} 
& \multicolumn{2}{c||}{ZIP} 
& \multicolumn{2}{c||}{FLARE} 
& \multicolumn{2}{c}{TRIM} \\
\cline{3-12}
& & CA$\uparrow$ & ASR$\downarrow$
& CA$\uparrow$ & ASR$\downarrow$
& CA$\uparrow$ & ASR$\downarrow$
& CA$\uparrow$ & ASR$\downarrow$
& CA$\uparrow$ & ASR$\downarrow$ \\
\hline\hline

\multirow{3}{*}{A2O}
& BadNets  & 85.12 & 96.03 & 72.15 & 8.43  & 67.67 & 13.23 & \textbf{81.20} & 19.45 & 78.68 & \textbf{5.98} \\
& Blended  & 83.26 & 99.97 & 65.02 & 19.50 & 63.15 & 88.12 & 68.14 & 20.03 & \textbf{77.12} & \textbf{6.12} \\
& Physical & 79.75 & 88.64 & 77.09 & 76.15 & 61.19 & 20.03 & 16.33 & 6.37 & \textbf{78.37} & \textbf{5.18} \\

\cline{2-12}
& \textbf{Avg}
& \cellcolor{red!10}{\bf82.04} & 94.21
& 71.89 & 34.69
& 64.12 & 40.46
& 55.22 & 15.28
& \cellcolor{green!20}{\bf78.06} & \cellcolor{green!20}{\bf5.76} \\

\hline\hline

\multirow{3}{*}{A2A}
& BadNets  & 73.92 & 46.56 & 70.06 & 7.65  & 70.74 & 7.13  & 72.03 & 21.89 & \textbf{74.71} & \textbf{6.14} \\
& Blended  & 70.16 & 58.13 & 67.11 & 15.15 & 67.98 & 23.30 & \textbf{69.56} & 29.08 & 68.47 & \textbf{9.20} \\
& Physical & 74.76 & 48.65 & \textbf{74.98} & 39.69 & 73.45 & 30.87 & 25.37 & 13.43 & 73.69 & \textbf{5.35} \\

\cline{2-12}
& \textbf{Avg}
& \cellcolor{red!10}{\bf72.95} & 51.78
& 70.19 & 20.83
& 70.38 & 20.43
& 55.64 & 21.47
& \cellcolor{green!20}{\bf72.29} & \cellcolor{green!20}{\bf6.90} \\

\hline
\end{tabular}
}
\vspace{-3mm}
\label{tab:imagenet_defense_comparison}
\end{table}

\subsection{Robustness to Trigger Size}
\label{sec:trigger_size}

\begin{table}[!t]
\centering

\caption{Comparison of TRIM vs. baselines on CIFAR-10 dataset with  larger $7{\times}7$-pixel triggers. Bold indicates best performance ($\uparrow$ higher CA, $\downarrow$ lower ASR).  All values are percentages (\%).}
\small
\setlength{\tabcolsep}{7pt}
\resizebox{\linewidth}{!}{
\begin{tabular}{l|l|cc|cc|cc|cc|cc}
\toprule

\multirow{2}{*}{\textbf{Dataset}} 
& \multirow{2}{*}{\textbf{Attack}} 
& \multicolumn{2}{c|}{\textbf{No Defense}} 
& \multicolumn{2}{c|}{\textbf{ShrinkPad}} 
& \multicolumn{2}{c|}{\textbf{ZIP}} 
& \multicolumn{2}{c|}{\textbf{FLARE}} 
& \multicolumn{2}{c}{\textbf{TRIM (Ours)}} \\

\cmidrule(lr){3-4}
\cmidrule(lr){5-6}
\cmidrule(lr){7-8}
\cmidrule(lr){9-10}
\cmidrule(lr){11-12}

& 
& \textbf{CA} $\uparrow$ & \textbf{ASR} $\downarrow$
& \textbf{CA} $\uparrow$ & \textbf{ASR} $\downarrow$
& \textbf{CA} $\uparrow$ & \textbf{ASR} $\downarrow$
& \textbf{CA} $\uparrow$ & \textbf{ASR} $\downarrow$
& \textbf{CA} $\uparrow$ & \textbf{ASR} $\downarrow$ \\

\midrule

\multirow{3}{*}{\rotatebox{90}{CIFAR-10}}

& BadNets  
& 82.31 & 98.99 
& 70.37 & 9.42  
& 76.90 & 39.12 
& 49.83 & 8.12 
& \textbf{87.20} & \textbf{0.55} \\

& Physical
& 89.12 & 97.77 
& 79.76 & 12.10 
& 53.16 & 71.12 
& 29.60 & 4.90 
& \textbf{86.70} & \textbf{1.79} \\

& LC
& 78.79 & 94.17 
& 79.15 & 17.67 
& 76.18 & 81.56 
& 57.54 & 8.17 
& \textbf{82.56} & \textbf{0.87} \\

\cmidrule(lr){2-12}

& \textit{Average} 
& 83.41 & 96.98 
& 76.43 & 13.06 
& 68.75 & 63.93 
& 45.66 & 7.06 
& \cellcolor{green!20}{\bf 85.49} & \cellcolor{green!20}{\bf 1.07} \\
\bottomrule
\end{tabular}
}
 \vspace{-5pt}
\label{tab:defense_comparison2}

\end{table}

To examine sensitivity to trigger scale, we additionally evaluate
$7\times7$ triggers on CIFAR-10. Table~\ref{tab:defense_comparison2} shows
that \textit{TRIM} achieves an average ASR of only $1.07\%$, compared with
$13.06\%$, $63.93\%$, and $7.06\%$ for ShrinkPad, ZIP, and FLARE,
respectively. At the same time, \textit{TRIM} retains an average CA of
$85.49\%$.

For the Physical attack, for example, \textit{TRIM} reduces ASR to $1.79\%$
while retaining $86.70\%$ CA. Under LC, it obtains $0.87\%$ ASR and
$82.56\%$ CA. These results indicate that the region-based defense remains
effective when the trigger scale differs from the primary evaluation
configuration.

\vspace{-3mm}

\subsection{Generalization to Diverse Trigger Constructions}
\label{sec:diverse_triggers}

Table~\ref{tab:various_attacks} extends the evaluation beyond conventional
fixed patch triggers to multi-trigger, distributed, input-aware, and semantic
trigger constructions. Without defense, these attacks remain highly
effective, reaching ASRs of up to $98.12\%$. \textit{TRIM} substantially
reduces attack success across all evaluated settings.

For CIFAR-10, ASR decreases from $97.17\%$ to $3.30\%$ for a single trigger,
from $91.56\%$ to $3.45\%$ when two triggers coexist, and from $88.21\%$ to
$2.03\%$ for DBA. The corresponding FPRs remain between $2.92\%$ and
$5.16\%$. On ImageNet-10, \textit{TRIM} reduces VSSC ASR from $91.61\%$
to $5.13\%$ while maintaining $78.73\%$ CA.

Input-aware triggers constitute the most challenging evaluated setting:
although \textit{TRIM} reduces ASR from $98.12\%$ to $16.34\%$, the residual
ASR is higher than for the other attacks. This result exposes an important
limitation: highly sample-specific patterns are less likely to match cached
trigger representations and can be more difficult to isolate using
region-wise intervention. We therefore report this case explicitly rather
than claiming uniform robustness across all trigger-generation mechanisms.

The two-trigger result further shows that \textit{TRIM} can identify and
purify multiple spatially separated suspicious regions in many cases.
However, as discussed in Sec.~\ref{sec:trigger_detection}, region-wise
prediction-change verification may miss a weaker trigger when another
dominant trigger independently maintains the backdoored prediction. The
multi-trigger experiment should therefore be interpreted as empirical
robustness under the evaluated configuration rather than a guarantee for
arbitrary interacting triggers.

\vspace{2mm}
\begin{table*}[!t]
\centering
\caption{Comparison of \textsc{TRIM} with baseline defenses under diverse
backdoor attacks on CIFAR-10 and ImageNet-10.  Bold indicates best performance ($\uparrow$ higher CA, $\downarrow$ lower ASR,  $\downarrow$ lower FPR).  All values are percentages (\%).}
\label{tab:various_attacks}

\setlength{\tabcolsep}{2.2pt}
\renewcommand{\arraystretch}{1.08}
\footnotesize

\resizebox{\textwidth}{!}{
\begin{tabular}{
l|l|
c|c|
c|c|c|
c|c|c|
c|c|c|
c|c|c
}
\toprule
\textbf{Dataset} & \textbf{Attack}
& \multicolumn{2}{c|}{\textbf{No Defense}}
& \multicolumn{3}{c|}{\textbf{ShrinkPad}}
& \multicolumn{3}{c|}{\textbf{ZIP}}
& \multicolumn{3}{c|}{\textbf{FLARE}}
& \multicolumn{3}{c}{\textbf{TRIM (Ours)}} \\

&
& CA$\uparrow$ & ASR$\downarrow$
& CA$\uparrow$ & ASR$\downarrow$ & FPR$\downarrow$
& CA$\uparrow$ & ASR$\downarrow$ & FPR$\downarrow$
& CA$\uparrow$ & ASR$\downarrow$ & FPR$\downarrow$
& CA$\uparrow$ & ASR$\downarrow$ & FPR$\downarrow$ \\
\midrule

\multirow{3}{*}{CIFAR-10}
& Two Triggers
& 89.73 & 91.56
& 71.50 & 38.50 & 10.20
& 79.50 & 19.80 & 7.80
& 81.80 & 11.50 & 6.10
& \textbf{84.13} & \textbf{3.45} & \textbf{5.16} \\

& Distributed (DBA)
& 86.45 & 88.21
& 72.80 & 19.60 & 7.50
& 78.20 & 18.20 & 7.00
& 83.00 & 7.20 & 3.95
& \textbf{84.56} & \textbf{2.03} & \textbf{2.92} \\

& Input-Aware
& 90.56 & 98.12
& 69.90 & 47.30 & 11.50
& 77.60 & 31.40 & 9.30
& 83.50 & 21.60 & 6.90
& \textbf{86.03} & \textbf{16.34} & \textbf{5.06} \\

\midrule

ImageNet-10
& VSSC
& 84.17 & 91.61
& 68.50 & 58.90 & 11.00
& 66.90 & 41.20 & 9.40
& \textbf{80.20} & 35.60 & 8.20
& 78.73 & \textbf{5.13} & \textbf{6.19} \\

\bottomrule
\end{tabular}
}
\vspace{-3mm}
\end{table*}

\subsection{Sensitivity and Component Analysis}
\label{sec:sensitivity}

\noindent\textbf{Effect of Segment-Area Threshold.} Table~\ref{tab:segmentation} studies the effect of the segment-area threshold on the number of evaluated regions, computational cost, artifact ratio (AR),
and post-purification ASR. The results reveal a clear trade-off: restrictive thresholds reduce candidate regions and computational cost but may miss trigger-related regions, increasing residual ASR. In contrast, permissive thresholds process more regions and risk modifying benign semantic content.

\begin{wraptable}{r}{0.59\textwidth}
 \vspace{-4mm}
 \centering
 \caption{Effect of the segment-area threshold on the average number of segments per image, inference-time inpainting cost ($T$), AR, and post-purification ASR.}
\resizebox{\linewidth}{!}{
\begin{tabular}{c|cccc||cccc}
\toprule
\multirow{3}{*}{\rotatebox{90}{\shortstack{\textbf{Seg.}\\\textbf{Area}}}} 
& \multicolumn{4}{c||}{\textbf{CIFAR-10}}
& \multicolumn{4}{c}{\textbf{ImageNet-10}} \\

\cmidrule(lr){2-5} \cmidrule(lr){6-9}
& \textbf{Segs} & \textbf{T(ms)}& \textbf{AR} & \textbf{ASR}& \textbf{Segs} & \textbf{T(s)} & \textbf{AR} & \textbf{ASR} \\

\midrule
0.1 & \cellcolor{green!20}{\bf2.81} & \cellcolor{green!20}{\bf3.02} & \cellcolor{green!20}{\bf2.30} & 7.5 & \cellcolor{green!20}{\bf47.82} & 6.71 & 0.60 & 10.5 \\
\cellcolor{red!10}{\bf 0.2} & 3.71 & 3.68 & 2.77 & \textbf{2.1} & 51.01 & \cellcolor{green!20}{\bf7.17} & \cellcolor{green!20}{\bf1.00} & \textbf{3.2} \\
\cellcolor{red!10}{\bf0.3} & 4.24 & 4.06 & 3.18 & \cellcolor{green!20}{\bf2.0} & 52.70 & 7.40 & 1.60 & \cellcolor{green!20}{\bf3.0} \\
0.4 & 4.56 & 4.35 & 3.50 & 3.5 & 53.93 & 7.57 & 2.20 & 5.0 \\
0.5 & 4.83 & 4.62 & 3.75 & 4.8 & 54.74 & 7.68 & 2.70 & 6.2 \\
0.6 & 5.05 & 4.83 & 4.00 & 5.6 & 55.31 & 7.76 & 3.10 & 7.5 \\
0.7 & 5.20 & 5.01 & 4.20 & 6.3 & 55.81 & 7.83 & 3.50 & 8.3 \\
0.8 & 5.28 & 5.06 & 4.35 & 7.0 & 55.99 & 7.86 & 3.90 & 9.0 \\
0.9 & 5.34 & 5.12 & 4.50 & 7.8 & 56.13 & 7.88 & 4.20 & 9.8 \\
1.0 & 6.39 & 6.32 & 5.00 & 8.5 & 57.25 & 8.04 & 4.50 & 10.6 \\
\bottomrule
\end{tabular}}
 \vspace{-2mm}
\label{tab:segmentation}
\end{wraptable}
For both CIFAR-10 and ImageNet-10, ASR is minimized around thresholds of  $0.2$--$0.3$. We therefore set $0.25$ as the default threshold, balancing robustness and artifact preservation, and fix it across all remaining experiments without attack-specific tuning.


AR complements ASR by quantifying unintended changes to clean inputs:
a configuration with low ASR but high AR would indicate overly aggressive
purification, whereas low AR with high ASR would indicate insufficient
trigger removal. We assess reconstruction artifacts using SSIM
~\cite{wang2004image} and LPIPS~\cite{zhang2018unreasonable}, together with
manual inspection of flagged cases.

\begin{figure}[!t]
    \centering
    \begin{minipage}{0.245\linewidth}
        \centering
              \subfigure[BadNets.]{\includegraphics[width=\linewidth]{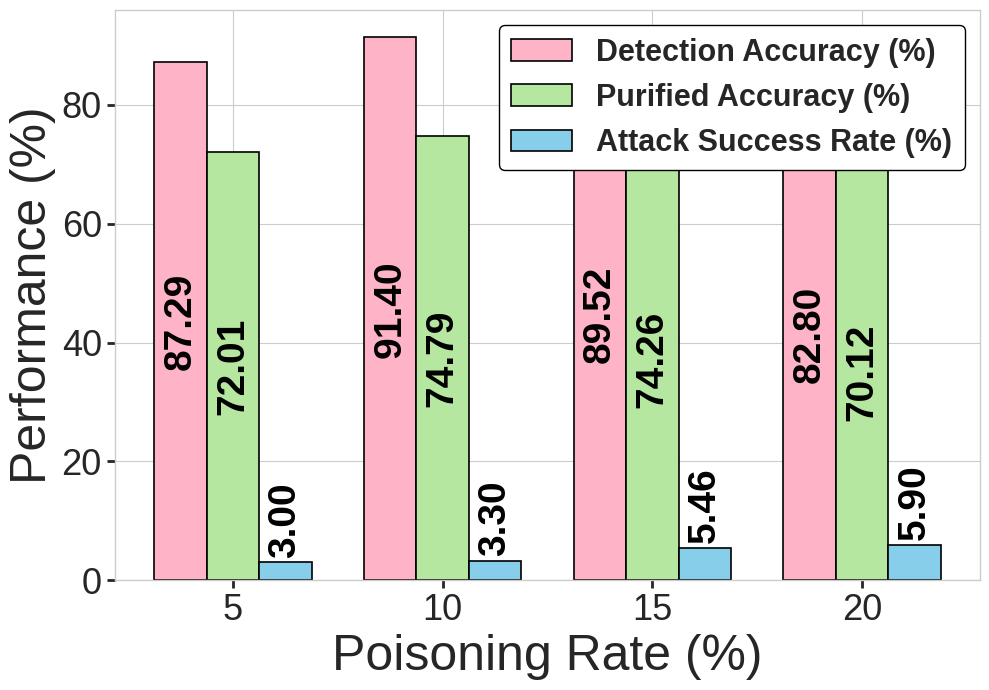}}
    \end{minipage}
    \begin{minipage}{0.245\linewidth}
        \centering
            \subfigure[Physical.]{\includegraphics[width=\linewidth]{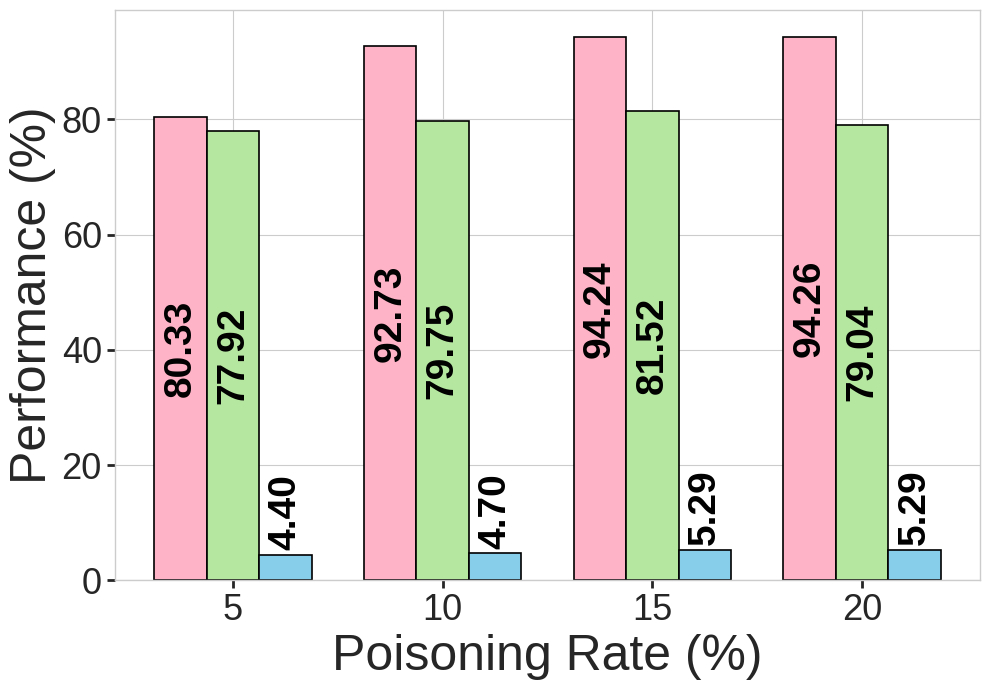}}
    \end{minipage}
    \begin{minipage}{0.245\linewidth}
        \centering
         \subfigure[LC.]{\includegraphics[width=\linewidth]{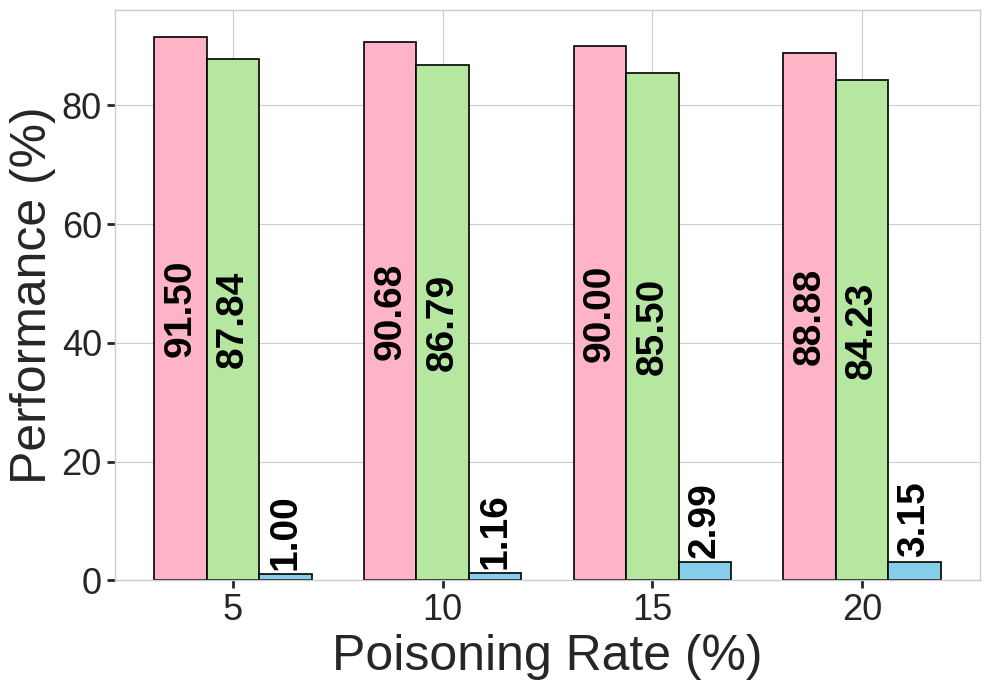}}
    \end{minipage}
    \begin{minipage}{0.245\linewidth}
        \centering
        \subfigure[Dynamic.]{\includegraphics[width=\linewidth]{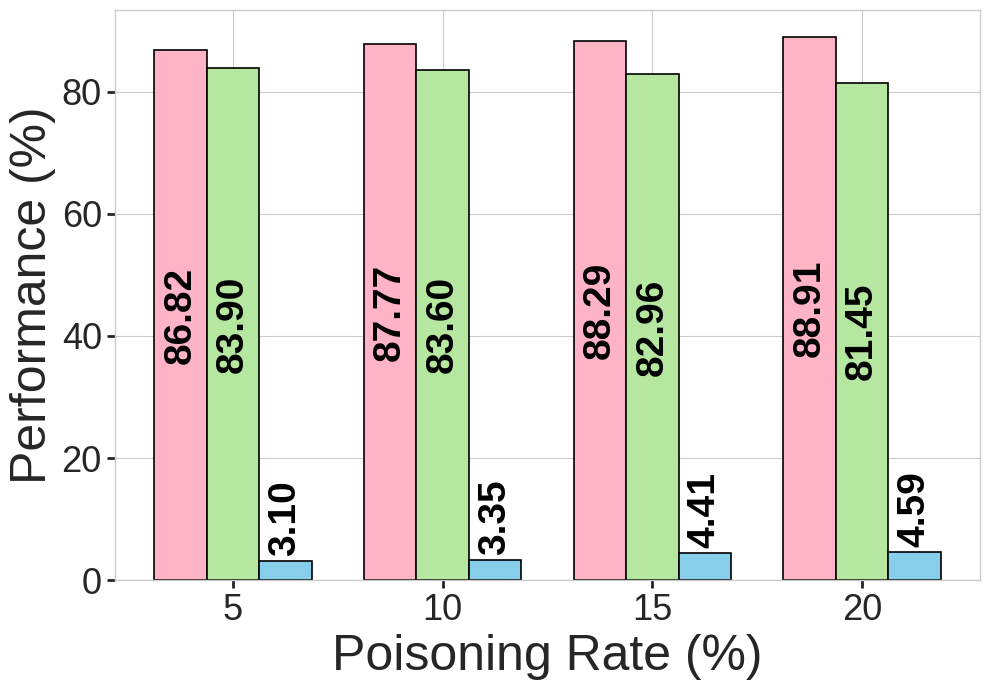}}
    \end{minipage}    
    \vspace{-2mm}
\caption{TRIM performance under poisoning rates of 5\%--20\% on CIFAR-10 and ImageNet-10, reporting DA, PA, and ASR for BadNets, Physical, LC, and Dynamic attacks.}
    \label{fig:poison_rate_effectiveness}

    \vspace{-3mm}
\end{figure}

 \vspace{2mm}
\noindent\textbf{Robustness Across Poisoning Rates.} \Cref{fig:poison_rate_effectiveness} evaluates \textit{TRIM} across poisoning rates of $5\%$--$20\%$. Across BadNets, Physical, LC, and Dynamic attacks, DA and PA remain stable, while ASR stays below $6\%$.


For BadNets, DA remains above $82\%$ and ASR ranges from $3.0\%$ to $5.9\%$. For Physical attacks, DA ranges from approximately $80\%$ to $94\%$, while ASR remains between $4.4\%$ and $5.3\%$. LC yields ASR between approximately $1\%$ and $3.35\%$, and Dynamic attacks remain between $3.10\%$ and $4.60\%$. These results indicate that \textit{TRIM}'s effectiveness is relatively stable over the evaluated poisoning-rate range rather than being tied to the default $10\%$ training-poisoning configuration.

\noindent\textbf{Effectiveness of Feature Caching.} Table~\ref{tab:inpainted_segments_reduction} quantifies

\begin{wraptable}{r}{0.38\textwidth}
 \vspace{-8.5mm}
\centering 
\caption{Percentage reduction of inpainted segments as the number of processed images increases.}
\resizebox{\linewidth}{!}{
\begin{tabular}{c|c|c}
\toprule
\textbf{Images Proc.} & \textbf{CIFAR-10} & \textbf{ImageNet-10} \\
\midrule
0     & 100.00 & 100.00 \\
50    & 78.54  & 93.16  \\
100   & 37.12  & 75.78  \\
250   & 31.50  & 69.49  \\
500   & 17.23  & 49.98  \\
750   & 11.05  & 32.11  \\
1000  & 10.23  & 27.66  \\
\bottomrule
\end{tabular}}
\label{tab:inpainted_segments_reduction}\vspace{-2mm}
\end{wraptable}
\noindent the computational benefit of RBF-based feature caching. At initialization, all candidate regions require analysis. As the cache accumulates previously verified benign and
trigger-related representations, the fraction of regions requiring inpainting
decreases substantially. After 1,000 processed images, only $10.23\%$ of
candidate regions on CIFAR-10 and $27.66\%$ on ImageNet-10 require
inpainting, compared with $100\%$ initially. This result provides direct evidence for the role of feature caching in reducing repeated reconstruction and black-box classifier queries. The larger
remaining fraction on ImageNet-10 is consistent with its greater visual
diversity, which yields fewer repeated region representations than CIFAR-10.

\subsection{Computational and Deployment Analysis}
\label{sec:computational}

Table~\ref{tab:TRIM_cost_breakdown} reports the per-component runtime, peak
VRAM usage, and FLOPs of \textit{TRIM}. The results clarify that computational
cost depends strongly on the selected segmentation and reconstruction
modules.

SAM ViT-H requires $0.53$\,s and approximately $7.00$\,GB of peak VRAM per
ImageNet-10-scale input, whereas MobileSAM reduces segmentation time to
$0.022$\,s with $0.08$\,GB. Feature extraction and RBF matching introduce
comparatively small overhead. The dominant cost in the high-fidelity
configuration is Stable Diffusion inpainting, requiring approximately
$5.87$\,s per reconstruction and $4.85$\,GB of VRAM. In contrast,
\texttt{cv2.inpaint} requires approximately $0.020$\,s.

These measurements show that \textit{TRIM} should not be characterized as
universally real-time when diffusion-based reconstruction is enabled.
Instead, the framework provides a configurable robustness--latency trade-off:
SAM ViT-H and diffusion inpainting prioritize segmentation and reconstruction
quality, whereas MobileSAM and classical inpainting provide a substantially
lighter deployment configuration. Feature caching further reduces amortized
cost by decreasing the number of candidate regions requiring reconstruction
as more inputs are processed.

\begin{table}[t]
\centering
\caption{Per-image runtime, peak VRAM usage, and computational cost (FLOPs) of major components in the proposed \textsc{TRIM} defense pipeline on ImageNet-10-scale inputs. Results are reported on a single GPU.}
\label{tab:TRIM_cost_breakdown}
\renewcommand{\arraystretch}{1}
\setlength{\tabcolsep}{10pt}
\footnotesize
\resizebox{\linewidth}{!}{
\begin{tabular}{llccc}
\hline
\textbf{Category} & \textbf{Component} & \textbf{Time (s)} & \textbf{VRAM (GB)} & \textbf{FLOPs (G)} \\
\hline
\multirow{2}{*}{Segmentation}
& SAM ViT-H Segmentation
& 0.53
& 7.00
& $2.7 \times 10^{3}$ \\
& MobileSAM
& 0.022
& 0.08
& $3.9 \times 10^{1}$ \\
\hline
\multirow{3}{*}{Feature Extraction}
& ResNet-18 Feature Extraction
& 0.02
& 0.04
& $1.8$ \\
& ResNet-50 Feature Extraction
& 0.04
& 0.10
& $4.1$ \\
& RBF Similarity Matching
& 0.001
& $<0.01$
& $5\times10^{-3}$ \\
\hline
\multirow{2}{*}{Inpainting}
& Stable Diffusion 2 Inpainting
& 5.87
& 4.85
& $1.7\times10^{4}$ \\
& cv2.inpaint
& 0.020
& 0
& $5\times10^{-3}$ \\
\hline
\multirow{2}{*}{Inference}
& Classifier Query (ResNet-18)
& 0.02
& 0.04
& $1.8$ \\
& Classifier Query (ResNet-50)
& 0.04
& 0.10
& $4.1$ \\
\hline
I/O \& Masking
& I/O + Mask Post-processing
& 0.003
& 0
& -- \\
\hline
\end{tabular}}
 \vspace{-4mm}
\end{table}

\vspace{-4mm}
\section{Conclusion}
We presented TRIM, a black-box inference-time defense that detects and mitigates backdoor attacks without access to model parameters, training data, or clean validation sets. By combining semantic segmentation, segment-level features, and selective inpainting, TRIM localizes and removes trigger regions to restore correct predictions. Experiments on CIFAR-10 and ImageNet-10 show that TRIM significantly reduces ASR—from 97.94\% to 3.77\% and from 78.45\% to 5.06\%, respectively—while maintaining high clean accuracy. Compared with ShrinkPad, ZIP, and FLARE, TRIM provides stronger protection against diverse triggers, including small, large, blended, and spatially sparse patterns.

 \bibliography{egbib}

@inproceedings{bagdasaryan2020backdoor,
  title={How to backdoor federated learning},
  author={Bagdasaryan, Eugene and Veit, Andreas and Hua, Yiqing and Estrin, Deborah and Shmatikov, Vitaly},
  booktitle={International conference on artificial intelligence and statistics},
  pages={2938--2948},
  year={2020},
  organization={PMLR}
}

@article{zhu2024breaking,
  title={Breaking the false sense of security in backdoor defense through re-activation attack},
  author={Zhu, Mingli and Liang, Siyuan and Wu, Baoyuan},
  journal={Advances in Neural Information Processing Systems},
  volume={37},
  pages={114928--114964},
  year={2024}
}

@inproceedings{salem2022dynamic,
  title={Dynamic backdoor attacks against machine learning models},
  author={Salem, Ahmed and Wen, Rui and Backes, Michael and Ma, Shiqing and Zhang, Yang},
  booktitle={2022 IEEE 7th European Symposium on Security and Privacy (EuroS\&P)},
  pages={703--718},
  year={2022},
  organization={IEEE}
}

@inproceedings{barni2019new,
  title={A new backdoor attack in cnns by training set corruption without label poisoning},
  author={Barni, Mauro and Kallas, Kassem and Tondi, Benedetta},
  booktitle={2019 IEEE International Conference on Image Processing (ICIP)},
  pages={101--105},
  year={2019},
  organization={IEEE}
}

@article{wang2021backdoor,
  title={Backdoor attack through frequency domain},
  author={Wang, Tong and Yao, Yuan and Xu, Feng and An, Shengwei and Tong, Hanghang and Wang, Ting},
  journal={arXiv preprint arXiv:2111.10991},
  year={2021}
}

@article{chen2018detecting,
  title={Detecting backdoor attacks on deep neural networks by activation clustering},
  author={Chen, Bryant and Carvalho, Wilka and Baracaldo, Nathalie and Ludwig, Heiko and Edwards, Benjamin and Lee, Taesung and Molloy, Ian and Srivastava, Biplav},
  journal={arXiv preprint arXiv:1811.03728},
  year={2018}
}

@inproceedings{hayase2021spectre,
  title={Spectre: Defending against backdoor attacks using robust statistics},
  author={Hayase, Jonathan and Kong, Weihao and Somani, Raghav and Oh, Sewoong},
  booktitle={International Conference on Machine Learning},
  pages={4129--4139},
  year={2021},
  organization={PMLR}
}

@article{zhu2023vdc,
  title={Vdc: Versatile data cleanser based on visual-linguistic inconsistency by multimodal large language models},
  author={Zhu, Zihao and Zhang, Mingda and Wei, Shaokui and Wu, Bingzhe and Wu, Baoyuan},
  journal={arXiv preprint arXiv:2309.16211},
  year={2023}
}

@inproceedings{gao2023backdoor,
  title={Backdoor defense via adaptively splitting poisoned dataset},
  author={Gao, Kuofeng and Bai, Yang and Gu, Jindong and Yang, Yong and Xia, Shu-Tao},
  booktitle={Proceedings of the IEEE/CVF Conference on Computer Vision and Pattern Recognition},
  pages={4005--4014},
  year={2023}
}

@article{chen2022effective,
  title={Effective backdoor defense by exploiting sensitivity of poisoned samples},
  author={Chen, Weixin and Wu, Baoyuan and Wang, Haoqian},
  journal={Advances in Neural Information Processing Systems},
  volume={35},
  pages={9727--9737},
  year={2022}
}

@article{li2021neural,
  title={Neural attention distillation: Erasing backdoor triggers from deep neural networks},
  author={Li, Yige and Lyu, Xixiang and Koren, Nodens and Lyu, Lingjuan and Li, Bo and Ma, Xingjun},
  journal={arXiv preprint arXiv:2101.05930},
  year={2021}
}

@article{guo2023scale,
  title={Scale-up: An efficient black-box input-level backdoor detection via analyzing scaled prediction consistency},
  author={Guo, Junfeng and Li, Yiming and Chen, Xun and Guo, Hanqing and Sun, Lichao and Liu, Cong},
  journal={arXiv preprint arXiv:2302.03251},
  year={2023}
}

@article{yang2024sampdetox,
  title={Sampdetox: Black-box backdoor defense via perturbation-based sample detoxification},
  author={Yang, Yanxin and Jia, Chentao and Yan, DengKe and Hu, Ming and Li, Tianlin and Xie, Xiaofei and Wei, Xian and Chen, Mingsong},
  journal={Advances in Neural Information Processing Systems},
  volume={37},
  pages={121236--121264},
  year={2024}
}

@article{shao2021bddr,
  title={Bddr: An effective defense against textual backdoor attacks},
  author={Shao, Kun and Yang, Junan and Ai, Yang and Liu, Hui and Zhang, Yu},
  journal={Computers \& Security},
  volume={110},
  pages={102433},
  year={2021},
  publisher={Elsevier}
}

@inproceedings{liu2023detecting,
  title={Detecting backdoors during the inference stage based on corruption robustness consistency},
  author={Liu, Xiaogeng and Li, Minghui and Wang, Haoyu and Hu, Shengshan and Ye, Dengpan and Jin, Hai and Wu, Libing and Xiao, Chaowei},
  booktitle={Proceedings of the IEEE/CVF Conference on Computer Vision and Pattern Recognition},
  pages={16363--16372},
  year={2023}
}

@inproceedings{huang2023orion,
  title={Orion: Online backdoor sample detection via evolution deviance.},
  author={Huang, Huayang and Wang, Qian and Gong, Xueluan and Wang, Tao},
  booktitle={IJCAI},
  pages={864--874},
  year={2023}
}

@article{shi2023black,
  title={Black-box backdoor defense via zero-shot image purification},
  author={Shi, Yucheng and Du, Mengnan and Wu, Xuansheng and Guan, Zihan and Sun, Jin and Liu, Ninghao},
  journal={Advances in Neural Information Processing Systems},
  volume={36},
  pages={57336--57366},
  year={2023}
}

@article{chen2017targeted,
  title={Targeted backdoor attacks on deep learning systems using data poisoning},
  author={Chen, Xinyun and Liu, Chang and Li, Bo and Lu, Kimberly and Song, Dawn},
  journal={arXiv preprint arXiv:1712.05526},
  year={2017}
}

@inproceedings{zhang2018unreasonable,
  title={The unreasonable effectiveness of deep features as a perceptual metric},
  author={Zhang, Richard and Isola, Phillip and Efros, Alexei A and Shechtman, Eli and Wang, Oliver},
  booktitle={Proceedings of the IEEE conference on computer vision and pattern recognition},
  pages={586--595},
  year={2018}
}

@article{wang2004image,
  title={Image quality assessment: from error visibility to structural similarity},
  author={Wang, Zhou and Bovik, Alan C and Sheikh, Hamid R and Simoncelli, Eero P},
  journal={IEEE transactions on image processing},
  volume={13},
  number={4},
  pages={600--612},
  year={2004},
  publisher={IEEE}
}

@article{li2022backdoor,
  title={Backdoor learning: A survey},
  author={Li, Yiming and Jiang, Yong and Li, Zhifeng and Xia, Shu-Tao},
  journal={IEEE transactions on neural networks and learning systems},
  volume={35},
  number={1},
  pages={5--22},
  year={2022},
  publisher={IEEE}
}

@inproceedings{liu2018trojaning,
  title={Trojaning attack on neural networks},
  author={Liu, Yingqi and Ma, Shiqing and Aafer, Yousra and Lee, Wen-Chuan and Zhai, Juan and Wang, Weihang and Zhang, Xiangyu},
  booktitle={25th Annual Network And Distributed System Security Symposium (NDSS 2018)},
  year={2018},
  organization={Internet Soc}
}

@article{doan2021backdoor,
  title={Backdoor attack with imperceptible input and latent modification},
  author={Doan, Khoa and Lao, Yingjie and Li, Ping},
  journal={Advances in Neural Information Processing Systems},
  volume={34},
  pages={18944--18957},
  year={2021}
}

@article{gao2020backdoor,
  title={Backdoor attacks and countermeasures on deep learning: A comprehensive review},
  author={Gao, Yansong and Doan, Bao Gia and Zhang, Zhi and Ma, Siqi and Zhang, Jiliang and Fu, Anmin and Nepal, Surya and Kim, Hyoungshick},
  journal={arXiv preprint arXiv:2007.10760},
  year={2020}
}

@article{li2021anti,
  title={Anti-backdoor learning: Training clean models on poisoned data},
  author={Li, Yige and Lyu, Xixiang and Koren, Nodens and Lyu, Lingjuan and Li, Bo and Ma, Xingjun},
  journal={Advances in Neural Information Processing Systems},
  volume={34},
  pages={14900--14912},
  year={2021}
}

@article{turner2019label,
  title={Label-consistent backdoor attacks},
  author={Turner, Alexander and Tsipras, Dimitris and Madry, Aleksander},
  journal={arXiv preprint arXiv:1912.02771},
  year={2019}
}

@article{li2020invisible,
  title={Invisible backdoor attacks on deep neural networks via steganography and regularization},
  author={Li, Shaofeng and Xue, Minhui and Zhao, Benjamin Zi Hao and Zhu, Haojin and Zhang, Xinpeng},
  journal={IEEE Transactions on Dependable and Secure Computing},
  volume={18},
  number={5},
  pages={2088--2105},
  year={2020},
  publisher={IEEE}
}

@inproceedings{kirillov2023segment,
  title={Segment anything},
  author={Kirillov, Alexander and Mintun, Eric and Ravi, Nikhila and Mao, Hanzi and Rolland, Chloe and Gustafson, Laura and Xiao, Tete and Whitehead, Spencer and Berg, Alexander C and Lo, Wan-Yen and others},
  booktitle={Proceedings of the IEEE/CVF international conference on computer vision},
  pages={4015--4026},
  year={2023}
}

@article{gu2017badnets,
  title={Badnets: Identifying vulnerabilities in the machine learning model supply chain},
  author={Gu, Tianyu and Dolan-Gavitt, Brendan and Garg, Siddharth},
  journal={arXiv preprint arXiv:1708.06733},
  year={2017}
}

@article{krizhevsky2009learning,
  title={Learning multiple layers of features from tiny images},
  author={Krizhevsky, Alex and Hinton, Geoffrey and others},
  year={2009},
  publisher={Toronto, ON, Canada}
}

@inproceedings{fei2004learning,
  title={Learning generative visual models from few training examples: An incremental bayesian approach tested on 101 object categories},
  author={Fei-Fei, Li and Fergus, Rob and Perona, Pietro},
  booktitle={2004 conference on computer vision and pattern recognition workshop},
  pages={178--178},
  year={2004},
  organization={IEEE}
}

@article{westerhoff2025scam,
  title={Scam: A real-world typographic robustness evaluation for multimodal foundation models},
  author={Westerhoff, Justus and Purelku, Erblina and Hackstein, Jakob and Loos, Jonas and Pinetzki, Leo and Rodner, Erik and Hufe, Lorenz},
  journal={arXiv preprint arXiv:2504.04893},
  year={2025}
}

@inproceedings{deng2009imagenet,
  title={Imagenet: A large-scale hierarchical image database},
  author={Deng, Jia and Dong, Wei and Socher, Richard and Li, Li-Jia and Li, Kai and Fei-Fei, Li},
  booktitle={2009 IEEE conference on computer vision and pattern recognition},
  pages={248--255},
  year={2009},
  organization={Ieee}
}

@misc{imagenette2020,
  author = {Howard, Jeremy},
  title = {Imagenette: A smaller subset of 10 easily classified classes from Imagenet},
  year = {2020},
  howpublished = {\url{https://github.com/fastai/imagenette}}
}

@inproceedings{guan2024backdoor,
  title={Backdoor defense via test-time detecting and repairing},
  author={Guan, Jiyang and Liang, Jian and He, Ran},
  booktitle={Proceedings of the IEEE/CVF Conference on Computer Vision and Pattern Recognition},
  pages={24564--24573},
  year={2024}
}

@inproceedings{zhang2023backdoor,
  title={Backdoor defense via deconfounded representation learning},
  author={Zhang, Zaixi and Liu, Qi and Wang, Zhicai and Lu, Zepu and Hu, Qingyong},
  booktitle={Proceedings of the IEEE/CVF Conference on Computer Vision and Pattern Recognition},
  pages={12228--12238},
  year={2023}
}

@article{chen2025refine,
  title={Refine: Inversion-free backdoor defense via model reprogramming},
  author={Chen, Yukun and Shao, Shuo and Huang, Enhao and Li, Yiming and Chen, Pin-Yu and Qin, Zhan and Ren, Kui},
  journal={arXiv preprint arXiv:2502.18508},
  year={2025}
}

@article{hou2025flare,
  title={Flare: Towards universal dataset purification against backdoor attacks},
  author={Hou, Linshan and Luo, Wei and Hua, Zhongyun and Chen, Songhua and Zhang, Leo Yu and Li, Yiming},
  journal={IEEE Transactions on Information Forensics and Security},
  year={2025},
  publisher={IEEE}
}

@article{bai2024backdoor,
  title={Backdoor attack and defense on deep learning: A survey},
  author={Bai, Yang and Xing, Gaojie and Wu, Hongyan and Rao, Zhihong and Ma, Chuan and Wang, Shiping and Liu, Xiaolei and Zhou, Yimin and Tang, Jiajia and Huang, Kaijun and others},
  journal={IEEE Transactions on Computational Social Systems},
  volume={12},
  number={1},
  pages={404--434},
  year={2024},
  publisher={IEEE}
}

@article{ma2022beatrix,
  title={The" Beatrix''Resurrections: Robust Backdoor Detection via Gram Matrices},
  author={Ma, Wanlun and Wang, Derui and Sun, Ruoxi and Xue, Minhui and Wen, Sheng and Xiang, Yang},
  journal={arXiv preprint arXiv:2209.11715},
  year={2022}
}

@article{telea2004image,
  title={An image inpainting technique based on the fast marching method},
  author={Telea, Alexandru},
  journal={Journal of graphics tools},
  volume={9},
  number={1},
  pages={23--34},
  year={2004},
  publisher={Taylor \& Francis}
}

@inproceedings{rombach2022high,
  title={High-resolution image synthesis with latent diffusion models},
  author={Rombach, Robin and Blattmann, Andreas and Lorenz, Dominik and Esser, Patrick and Ommer, Bj{\"o}rn},
  booktitle={Proceedings of the IEEE/CVF conference on computer vision and pattern recognition},
  pages={10684--10695},
  year={2022}
}

@inproceedings{wenger2021backdoor,
  title={Backdoor attacks against deep learning systems in the physical world},
  author={Wenger, Emily and Passananti, Josephine and Bhagoji, Arjun Nitin and Yao, Yuanshun and Zheng, Haitao and Zhao, Ben Y},
  booktitle={Proceedings of the IEEE/CVF conference on computer vision and pattern recognition},
  pages={6206--6215},
  year={2021}
}

@article{xie2020dba,
  title={DBA: Distributed Backdoor Attacks against Federated Learning.},
  author={Xie, Chulin and Huang, Keli and Chen, Pin-Yu and Li, Bo},
  journal={ICLR},
  volume={2020},
  pages={1--19},
  year={2020}
}

@article{nguyen2020input,
  title={Input-aware dynamic backdoor attack},
  author={Nguyen, Tuan Anh and Tran, Anh},
  journal={Advances in Neural Information Processing Systems},
  volume={33},
  pages={3454--3464},
  year={2020}
}

@article{wang2023robust,
  title={Robust Backdoor Attack with Visible, Semantic, Sample-specific and Compatible Triggers},
  author={Wang, Ruotong and Chen, Hongrui and Zhu, Zihao and Liu, Li and Zhang, Yong and Fan, Yanbo and Wu, Baoyuan},
  year={2023}
}

@article{zhang2023faster,
  title={Faster segment anything: Towards lightweight sam for mobile applications},
  author={Zhang, Chaoning and Han, Dongshen and Qiao, Yu and Kim, Jung Uk and Bae, Sung-Ho and Lee, Seungkyu and Hong, Choong Seon},
  journal={arXiv preprint arXiv:2306.14289},
  year={2023}
}

@article{xue2026unified,
  title={A Unified Framework for Backdoor Trigger Segmentation},
  author={Xue, Dizhan and Qian, Shengsheng and Xu, Changsheng},
  journal={IEEE Transactions on Image Processing},
  year={2026},
  publisher={IEEE}
}

@inproceedings{radford2021learning,
  title={Learning transferable visual models from natural language supervision},
  author={Radford, Alec and Kim, Jong Wook and Hallacy, Chris and Ramesh, Aditya and Goh, Gabriel and Agarwal, Sandhini and Sastry, Girish and Askell, Amanda and Mishkin, Pamela and Clark, Jack and others},
  booktitle={International conference on machine learning},
  pages={8748--8763},
  year={2021},
  organization={PmLR}
}

@inproceedings{li2023blip,
  title={Blip-2: Bootstrapping language-image pre-training with frozen image encoders and large language models},
  author={Li, Junnan and Li, Dongxu and Savarese, Silvio and Hoi, Steven},
  booktitle={International conference on machine learning},
  pages={19730--19742},
  year={2023},
  organization={PMLR}
}

@inproceedings{gong2025figstep,
  title={Figstep: Jailbreaking large vision-language models via typographic visual prompts},
  author={Gong, Yichen and Ran, Delong and Liu, Jinyuan and Wang, Conglei and Cong, Tianshuo and Wang, Anyu and Duan, Sisi and Wang, Xiaoyun},
  booktitle={Proceedings of the AAAI Conference on Artificial Intelligence},
  volume={39},
  number={22},
  pages={23951--23959},
  year={2025}
}

@inproceedings{azuma2023defense,
  title={Defense-prefix for preventing typographic attacks on clip},
  author={Azuma, Hiroki and Matsui, Yusuke},
  booktitle={Proceedings of the IEEE/CVF International Conference on Computer Vision},
  pages={3644--3653},
  year={2023}
}

@inproceedings{stallkamp2011german,
  title={The German traffic sign recognition benchmark: a multi-class classification competition},
  author={Stallkamp, Johannes and Schlipsing, Marc and Salmen, Jan and Igel, Christian},
  booktitle={The 2011 international joint conference on neural networks},
  pages={1453--1460},
  year={2011},
  organization={IEEE}
}

@inproceedings{abdelnaby2026region,
  title={Region-Level Black-Box Defense Against Stealthy Embedding-Space Backdoors in CLIP},
  author={Abdelnaby, Ahmed and Elmahallawy, Mohamed},
  booktitle={Pacific-Asia Conference on Knowledge Discovery and Data Mining},
  pages={240--255},
  year={2026},
  organization={Springer}
}
\clearpage
\appendix
\section*{Appendix}
 
\section{Evaluation of TRIM on Additional Datasets}
 
To further demonstrate TRIM's robustness, generalizability, and purification effectiveness, we extend our evaluation to a broader set of datasets covering both low- and high-resolution settings. 
For low-resolution data, we use {\em CIFAR-100}~\cite{krizhevsky2009learning}, which contains 60{,}000 images of size $32{\times}32$ across 100 fine-grained classes, offering a more diverse and challenging classification problem than CIFAR-10. 
For high-resolution evaluation, we adopt {\em Caltech-101}~\cite{fei2004learning}, which contains around 9{,}000 images spanning 101 object categories with varied shapes, textures, and backgrounds; its higher resolution of $300{\times}200$ and intra-class variability introduce additional complexity for trigger localization and purification. We also include {\em GTSRB} ~\cite{stallkamp2011german} (German Traffic Sign Recognition Benchmark), consisting of over 50{,}000 traffic sign images with significant variations in illumination, viewpoint, occlusion, and resolutions ranging from $15{\times}15$ to $250{\times}250$ pixels  -- a setting where localized triggers can be especially difficult to distinguish from natural patterns. 
Together, these datasets allow us to assess TRIM’s performance across diverse visual domains, resolutions, and structural complexities.

In Table~\ref{tab:backdoor-defense}, we observe that TRIM consistently suppresses ASR across diverse datasets and attack types while preserving strong CA. Classical backdoor attacks such as \textit{BadNets} and \textit{Blended}, which typically achieve extremely high ASR (95--99\%), are almost completely neutralized by TRIM, with post-defense ASR reduced to 0.34--6\% in most settings. For more challenging scenarios, including \textit{Physical} attacks and higher-variability datasets such as GTSRB and Caltech-101, TRIM still achieves substantial reductions in the attack success rate. For instance, on GTSRB, ASR drops from 99.21\% to 4.22\%, and on Caltech-101, from 95.33\% to 2.26\%, demonstrating that TRIM remains effective even when triggers blend naturally with complex backgrounds or high-resolution content.

Notably, CA remains stable—and in some cases slightly improves—after purification, reaching 99.01\%, indicating that TRIM {\em avoids over-purification and preserves task-relevant visual features}. These results collectively highlight the robustness, generality, and scalability of TRIM across heterogeneous datasets, resolutions, and trigger designs.

\begin{table}[!t]
\centering
\small
\label{sec:ablation}
\setlength{\tabcolsep}{6pt}
 
\caption{Comparison of TRIM and the no-defense baseline on MNIST, CIFAR-100, GTSRB, and Caltech-101. We report clean accuracy (CA, $\uparrow$) and attack success rate (ASR, $\downarrow$) for three backdoor attacks (BadNets, Blended, and Physical).}
\resizebox{\linewidth}{!}{
\begin{tabular}{l l c c c c c c}
\toprule
& & \multicolumn{2}{c}{\textbf{BadNets}} 
  & \multicolumn{2}{c}{\textbf{Blended}} 
  & \multicolumn{2}{c}{\textbf{Physical}} \\
\cmidrule(lr){3-4} \cmidrule(lr){5-6} \cmidrule(lr){7-8}
\textbf{Dataset} & \textbf{Setting} 
& \textbf{CA} $\uparrow$ & \textbf{ASR} $\downarrow$
& \textbf{CA} $\uparrow$ & \textbf{ASR} $\downarrow$
& \textbf{CA} $\uparrow$ & \textbf{ASR} $\downarrow$ \\
\midrule

\multirow{2}{*}{\textbf{MNIST}}
& w/o defense  & 98.19 & 99.67 & 97.19 & 99.89 & 97.54 & 99.14 \\
& TRIM & \cellcolor{red!20} {\bf 99.01} & \cellcolor{green!20}0.34  & \cellcolor{red!20}95.24 &\cellcolor{green!20} 3.43  & \cellcolor{red!20}97.13 & \cellcolor{green!20}{\bf 0.23}  \\
\midrule

\multirow{2}{*}{\textbf{CIFAR-100}}
& w/o defense  & 76.15 & 99.49 & 73.34 & 99.90 & 70.15 & 98.40 \\
& TRIM & \cellcolor{red!20}77.43 & \cellcolor{green!20}6.36  & \cellcolor{red!20}71.21 & \cellcolor{green!20}10.08 & \cellcolor{red!20}68.14 & \cellcolor{green!20}5.32  \\
\midrule

\multirow{2}{*}{\textbf{GTSRB}}
& w/o defense  & 97.03 & 97.86 & 96.13 & 97.89 & 96.98 & 99.21 \\
& TRIM & \cellcolor{red!20}95.78 & \cellcolor{green!20}3.14  & \cellcolor{red!20}89.13 & \cellcolor{green!20}6.45  & \cellcolor{red!20}87.78 & \cellcolor{green!20}4.22  \\
\midrule

\multirow{2}{*}{\textbf{Caltech-101}}
& w/o defense  & 89.93 & 88.42 & 85.78 & 92.65 & 88.78 & 95.33 \\
& TRIM & \cellcolor{red!20}85.46 & \cellcolor{green!20}1.31  & \cellcolor{red!20}86.05 & \cellcolor{green!20}9.94  & \cellcolor{red!20}86.81 & \cellcolor{green!20}2.26  \\
\bottomrule
\end{tabular}}
\label{tab:backdoor-defense}
\end{table}

\section{Sensitivity Analysis}

Table~\ref{tab:sensitivity} presents a sensitivity analysis of several key hyperparameters used in  TRIM. Experiments are conducted on a subset of ImageNet-10 with a poisoning rate of $10\%$, where a physical patch trigger of size $24\times24$ pixels is applied. Additionally, in this analysis, SDI is used to reconstruct suspicious regions detected through segmentation and similarity filtering; however,  \texttt{cv2.inpaint} can also be used for faster processing.
The table reports CA, ASR without defense (w/o defense), ASR with defense (w/ defense), the average number of generated segments per image (Segs.), and the average defense time per image in seconds ($T_d$). The segmentation ratio controls the proportion of candidate regions selected for purification. Lower values (e.g., $0.15$) concentrate the defense on the most suspicious regions, preserving CA ($78.5\%$) while significantly reducing post-defense ASR to $3.6\%$. In contrast, larger ratios increase the number of processed segments and may degrade CA due to unnecessary modifications.

The similarity threshold determines when two segments are considered redundant based on the RBF feature similarity. Higher thresholds (e.g., $0.95$) avoid repeated purification of similar regions, improving efficiency while maintaining strong attack mitigation. The minimum SAM mask area  (i.e., 32) controls the smallest region retained after segmentation. Increasing this value filters out very small segments, reducing segmentation density and improving runtime, although overly large thresholds may remove useful region boundaries.

The mask merging IoU threshold determines when overlapping masks are combined. Moderate values (e.g., $0.5$) provide a balance between segmentation precision and region compactness, whereas extreme values may either fragment or over-merge regions. Finally, the number of SAM sampling points per side controls segmentation granularity: fewer points generate coarser masks with faster runtime, while larger values produce more detailed segmentation at the cost of higher computation.

Overall, the results indicate that the proposed framework remains robust across a wide range of parameter settings, while moderate segmentation ratios and similarity thresholds provide the best trade-off between clean accuracy preservation, attack mitigation, and computational efficiency. 

\begin{table}[!t]
\centering
\scriptsize

\caption{Sensitivity analysis of defense hyperparameters on an ImageNet-10 subset under $10\%$ poisoning with a $24\times24$ physical patch trigger. CA (\%), ASR w/o defense (\%), and ASR w/ defense (\%) denote clean accuracy and attack success rate before and after applying the defense. Segs. is the average number of segments and $T_d$ (s) is the runtime per image.}

\resizebox{\linewidth}{!}{
\begin{tabular}{l||cc||cccc}
\hline
\multirow{2}{*}{Experiment}
& \multicolumn{2}{c||}{Without Defense}
& \multicolumn{4}{c}{With Defense} \\
\cline{2-7}
& CA$\uparrow$ & ASR$\downarrow$
& CA$\uparrow$ & ASR$\downarrow$
& Segs.$\downarrow$ & $T_d$(s)$\downarrow$ \\
\hline\hline

\multicolumn{7}{c}{\textit{Segmentation Ratio}} \\
\hline
0.15 & 79.11 & 90.7 & \textbf{78.5} & \textbf{3.6} & 49.04 & 7.09 \\
0.35 & 79.11 & 90.7 & 62.0 & 9.3 & 51.83 & 7.19 \\
0.50 & 79.11 & 90.7 & 38.2 & 12.1 & 53.36 & 7.32 \\
0.75 & 79.11 & 90.7 & 37.4 & 12.4 & 53.69 & 7.36 \\

\hline\hline
\multicolumn{7}{c}{\textit{Similarity Threshold}} \\
\hline
0.80 & 79.11 & 90.7 & 24.9 & 11.8 & 51.83 & 6.89 \\
0.85 & 79.11 & 90.7 & 46.2 & 11.8 & 51.83 & 6.98 \\
0.90 & 79.11 & 90.7 & 62.0 & 9.3 & 51.83 & 7.19 \\
0.95 & 79.11 & 90.7 & 73.1 & 6.4 & 51.83 & 7.39 \\

\hline\hline
\multicolumn{7}{c}{\textit{SAM Minimum Area}} \\
\hline
32  & 79.11 & 90.7 & 62.0 & 9.3 & 51.83 & 7.21 \\
64  & 79.11 & 90.7 & 62.0 & 9.3 & 51.83 & 7.19 \\
128 & 79.11 & 90.7 & 60.3 & 9.8 & 48.84 & 2.14 \\

\hline\hline
\multicolumn{7}{c}{\textit{Mask Merge IoU}} \\
\hline
0.30 & 79.11 & 90.7 & 60.6 & 18.1 & 44.69 & 7.03 \\
0.50 & 79.11 & 90.7 & 62.0 & 9.3 & 51.83 & 7.20 \\
0.70 & 79.11 & 90.7 & 55.5 & 10.4 & 58.07 & 7.36 \\

\hline\hline
\multicolumn{7}{c}{\textit{SAM Points Per Side}} \\
\hline
16 & 79.11 & 90.7 & 64.3 & 9.5 & 31.37 & \textbf{6.05} \\
32 & 79.11 & 90.7 & 62.0 & 9.3 & 51.83 & 6.20 \\
64 & 79.11 & 90.7 & 62.6 & 8.7 & 61.54 & 10.36 \\

\hline
\end{tabular}
}

\label{tab:sensitivity}
\end{table}

\section{Comprehensive Comparison Across Attack Modes for ImageNet-10}
As shown in Table~\ref{tab:imagenet_defense_comparison}, 
the five evaluated defenses exhibit markedly different behaviors across both A2One and All2All 
attack settings. Without any defense, the models suffer from extremely high ASR values 
(\textit{e.g.}, $96.03\%$ for A2One BadNets and $99.97\%$ for A2One Blended), indicating that 
ImageNet-scale backdoor attacks remain highly effective on unprotected models. Traditional 
input-level defenses such as ShrinkPad and ZIP reduce the ASR to moderate levels 
(\textit{e.g.}, ShrinkPad achieves $8.43\%$ ASR for A2One BadNets and $7.65\%$ ASR in All2All), 
but these gains often come at the cost of reduced clean accuracy, with average CA dropping to 
$71.89\%$ and $70.19\%$ for A2One and A2A, respectively. 

FLARE shows mixed behavior: while it suppresses ASR in some cases (\textit{e.g.}, $6.37\%$ for 
A2One Physical), its average CA remains substantially lower, especially under complex 
different attack settings (\textit{e.g.}, A2One average CA of only $55.22\%$). In contrast, 
our TRIM defense consistently delivers the strongest robustness–utility trade-off. TRIM 
achieves the lowest average ASR across all settings ($5.76\%$ for A2One and $6.90\%$ for 
All2All) while simultaneously maintaining high clean accuracy comparable to the undefended 
model (\textit{e.g.}, $78.06\%$ CA for A2One and $72.29\%$ CA for All2All). These results 
demonstrate that TRIM effectively removes or neutralizes trigger regions without degrading 
semantic content, providing reliable resilience against diverse backdoor attacks on ImageNet-10.
\begin{table}[t!]
\centering
\scriptsize
\renewcommand{\arraystretch}{1.15}
\setlength{\tabcolsep}{2.5pt}

\caption{Comparison of TRIM vs. baselines on various attack modes for ImageNet-10. Bold indicates best performance ($\uparrow$ higher CA, $\downarrow$ lower ASR). All values are percentages.}

\resizebox{\columnwidth}{!}{
\begin{tabular}{l|c||cc||cc||cc||cc||cc}
\hline
\multirow{2}{*}{Attack} & \multirow{2}{*}{Method} 
& \multicolumn{2}{c||}{No Def.} 
& \multicolumn{2}{c||}{ShrinkPad} 
& \multicolumn{2}{c||}{ZIP} 
& \multicolumn{2}{c||}{FLARE} 
& \multicolumn{2}{c}{TRIM} \\
\cline{3-12}
& & CA$\uparrow$ & ASR$\downarrow$
& CA$\uparrow$ & ASR$\downarrow$
& CA$\uparrow$ & ASR$\downarrow$
& CA$\uparrow$ & ASR$\downarrow$
& CA$\uparrow$ & ASR$\downarrow$ \\
\hline\hline

\multirow{3}{*}{A2O}
& BadNets  & 85.12 & 96.03 & 72.15 & 8.43  & 67.67 & 13.23 & \textbf{81.20} & 19.45 & 78.68 & \textbf{5.98} \\
& Blended  & 83.26 & 99.97 & 65.02 & 19.50 & 63.50 & 88.12 & 68.14 & 20.03 & \textbf{77.12} & \textbf{6.12} \\
& Physical & 79.75 & 88.64 & 77.09 & 76.15 & 61.19 & 20.03 & 16.33 & 6.37 & \textbf{78.37} & \textbf{5.18} \\

\cline{2-12}
& \textbf{Avg}
& \cellcolor{red!10}{\bf82.04} & 94.21
& 71.89 & 34.69
& 64.12 & 40.46
& 55.22 & 15.28
& \cellcolor{green!20}{\bf78.06} & \cellcolor{green!20}{\bf5.76} \\

\hline\hline

\multirow{3}{*}{A2A}
& BadNets  & 73.92 & 46.56 & 70.06 & 7.65  & 70.74 & 7.13  & 72.03 & 21.89 & \textbf{74.71} & \textbf{6.14} \\
& Blended  & 70.16 & 58.13 & 67.11 & 15.15 & 67.98 & 23.30 & \textbf{69.56} & 29.08 & 68.47 & \textbf{9.20} \\
& Physical & 74.76 & 48.65 & \textbf{74.98} & 39.69 & 73.45 & 30.87 & 25.37 & 13.43 & 73.69 & \textbf{5.35} \\

\cline{2-12}
& \textbf{Avg}
& \cellcolor{red!10}{\bf72.95} & 51.78
& 70.19 & 20.83
& 70.38 & 20.43
& 55.64 & 21.47
& \cellcolor{green!20}{\bf72.29} & \cellcolor{green!20}{\bf6.90} \\

\hline
\end{tabular}
}

\label{tab:imagenet_defense_comparison}
\end{table}
\section{Grad-CAM Ablation: Trigger Effects and TRIM’s Response}

\begin{figure}[!t]
    \centering
    \includegraphics[width=\linewidth]{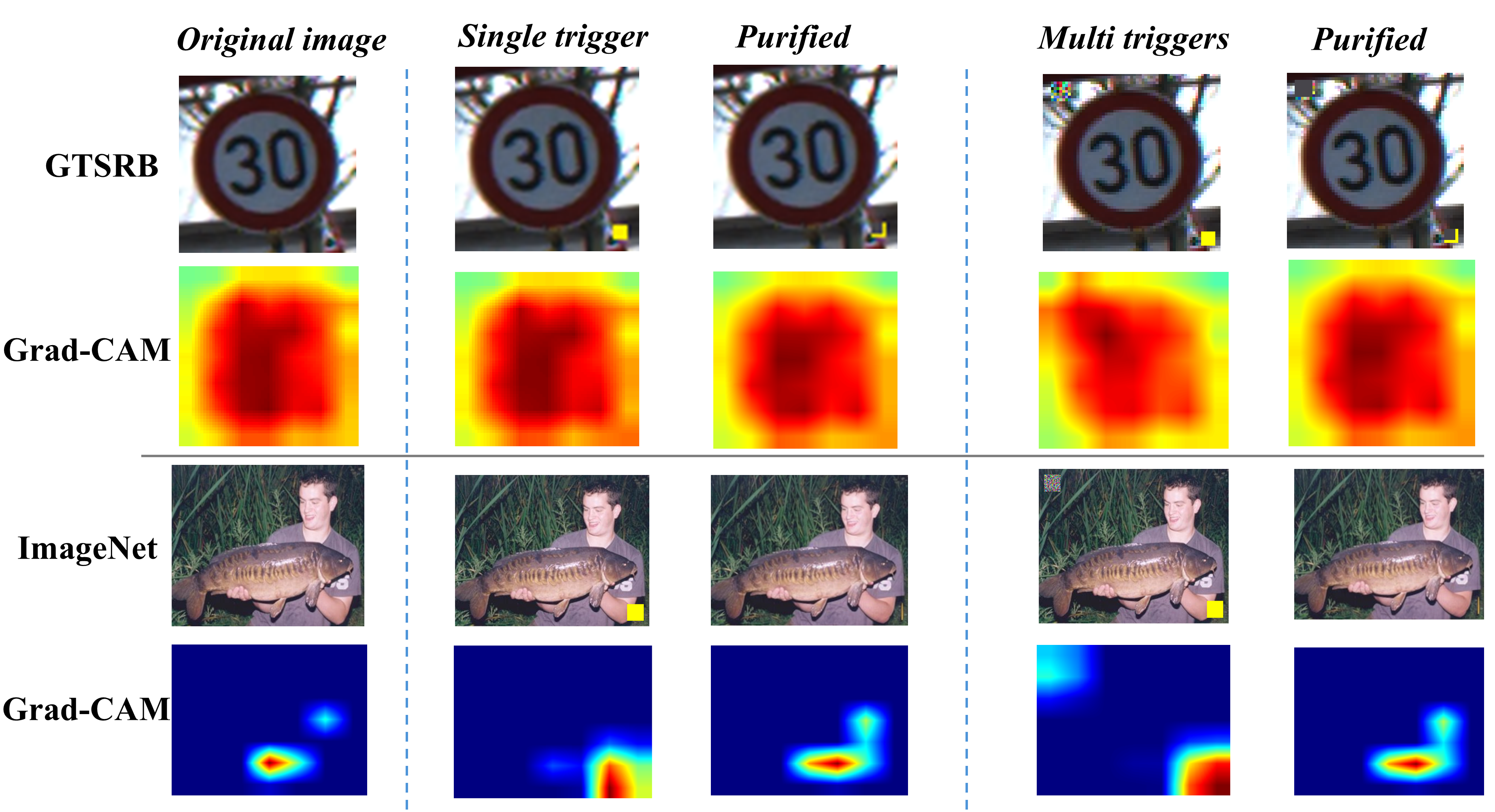}
\caption{Grad-CAM ablation for single- and multi-trigger backdoor attacks. \textbf{Top:} GTSRB samples on ResNet-18 with a BadNets yellow square (class~1 $\rightarrow$ class~2) and a physical-style patch (class~1 $\rightarrow$ class~5). \textbf{Bottom:} ImageNet-10 samples on ResNet-50 with BadNets (Tench $\rightarrow$ Cassette Player) and physical (Tench $\rightarrow$ Gas Pump) triggers. Clean images exhibit coherent, object-centered attention, while backdoored images show localized or dispersed saliency shifts around injected patches. TRIM removes the dominant trigger in a single purification pass, realigning model attention with the true semantic object.}

    \label{fig:gradcam_ablation}
\end{figure}

To better understand how different trigger types influence model behavior—and how TRIM's purification rectifies these effects—we conduct a Grad-CAM ablation study on both GTSRB and ImageNet-10 samples (Fig.~\ref{fig:gradcam_ablation}). For GTSRB, we evaluate a ResNet-18 classifier under two settings: (i) a \emph{single-trigger} BadNets yellow square that flips \textit{Speed Limit 30} (class~1) to \textit{Speed Limit 50} (class~2), and (ii) a \emph{multi-trigger} configuration that combines the same BadNets patch with a small physical-style noise square placed in the top-left corner, redirecting class~1 to \textit{Speed Limit 80} (class~5).

In a realistic deployment, TRIM performs only \emph{one purification pass per image}, removing the first detected trigger. Handling multiple triggers would require purifying and evaluating multiple image variants, which would increase inference time. As a result, in the two-trigger scenario, the BadNets patch is consistently detected first—demonstrating that certain triggers exert stronger influence than others and that detection naturally proceeds from dominant to weaker triggers across test-time samples. A parallel evaluation on ImageNet-10 using ResNet-50 follows the same structure: a BadNets patch flips \textit{Tench} to \textit{Cassette Player}, and in the multi-trigger setting, a physical patch induces \textit{Tench} to \textit{Gas Pump}.

Across both datasets, Grad-CAM visualizations reveal a consistent pattern: clean images show focused, object-centric attention; single- and multi-trigger images exhibit localized or dispersed saliency around the injected patches; and purified images restore attention to the true semantic object—returning to the center of the traffic sign in GTSRB and to the fish in ImageNet-10. These results confirm that TRIM reliably suppresses trigger-induced distortions and reinstates clean, human-aligned model reasoning, even under heterogeneous and stronger multi-trigger attacks.

\begin{figure*}[!t]
\centering

\subfigure[Poisoned Image.]{
    \includegraphics[width=0.23\linewidth]{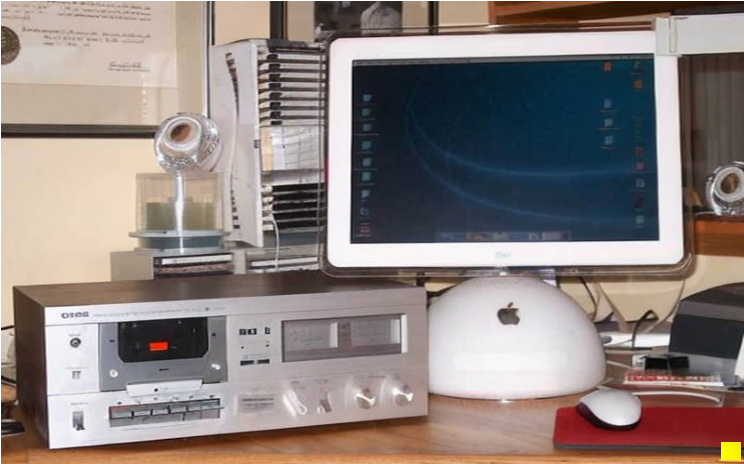}
}
\hfill
\subfigure[Cassette player door removed (Class not changed).]{
    \includegraphics[width=0.23\linewidth]{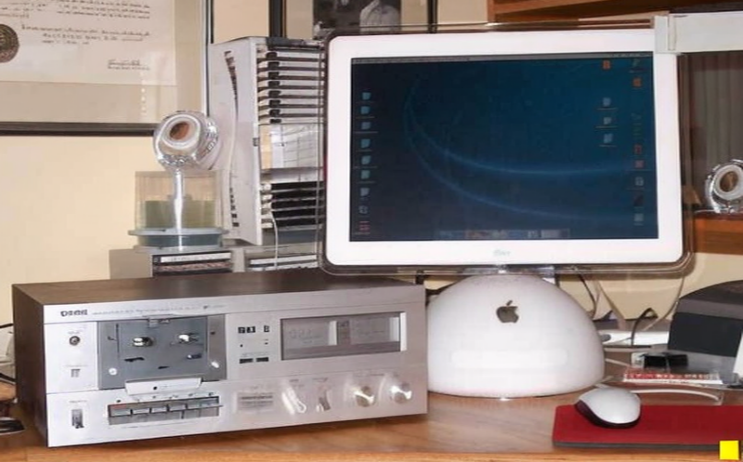}
}
\hfill
\subfigure[Red piece removed (Class not changed).]{
    \includegraphics[width=0.23\linewidth]{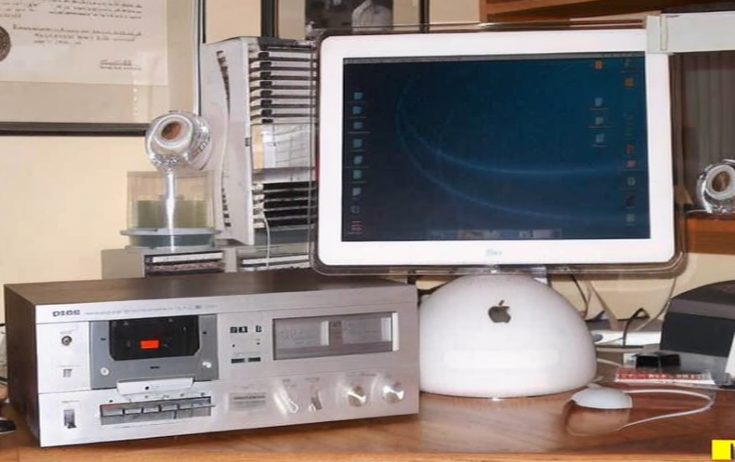}
}
\hfill
\subfigure[Black box at bottom right removed (Class not changed).]{
    \includegraphics[width=0.23\linewidth]{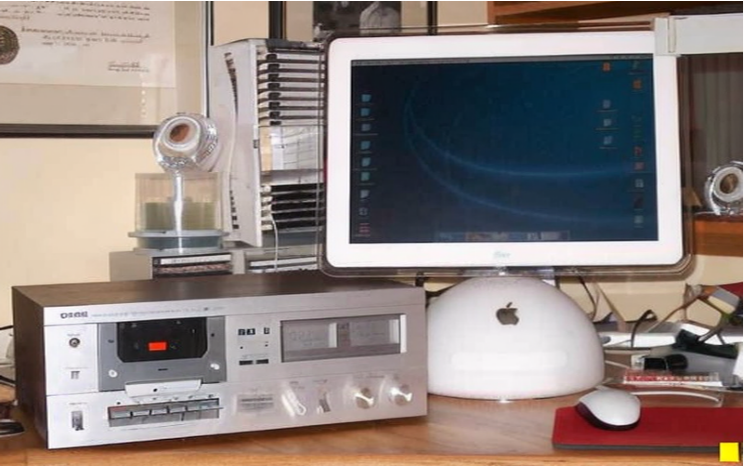}
}

\vspace{2mm}

\subfigure[Mouse removed (Class not changed).]{
    \includegraphics[width=0.23\linewidth]{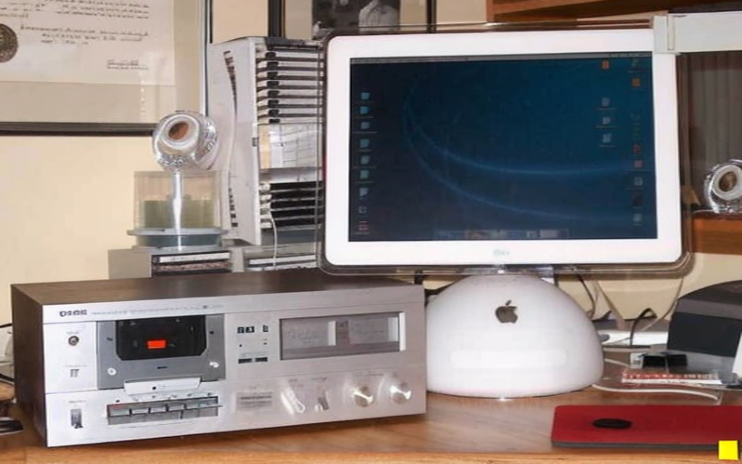}
}
\hfill
\subfigure[Paper removed (Class not changed).]{
    \includegraphics[width=0.23\linewidth]{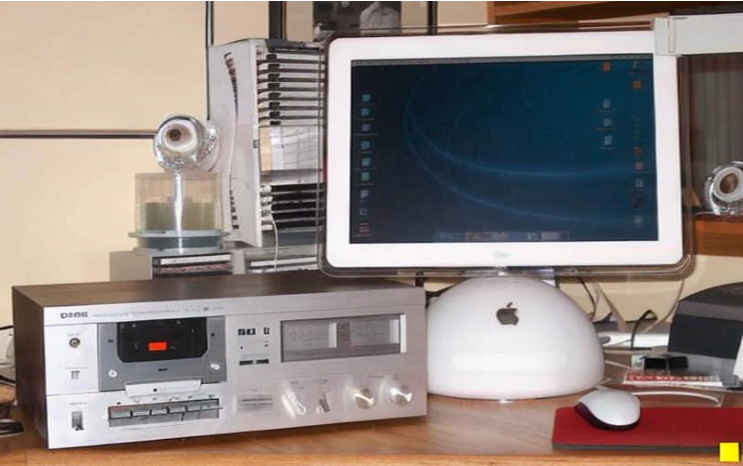}
}
\hfill
\subfigure[Base of PC monitor removed (Class not changed).]{
    \includegraphics[width=0.23\linewidth]{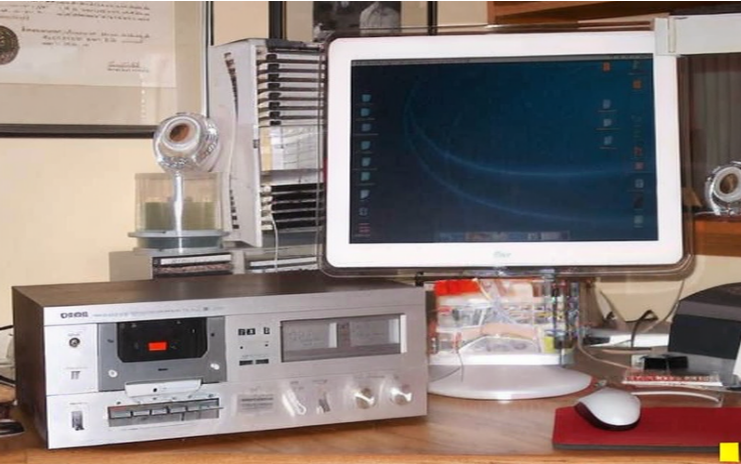}
}
\hfill
\subfigure[Yellow square trigger removed (\textbf{Class changed to cassette player}).]{
    \includegraphics[width=0.23\linewidth]{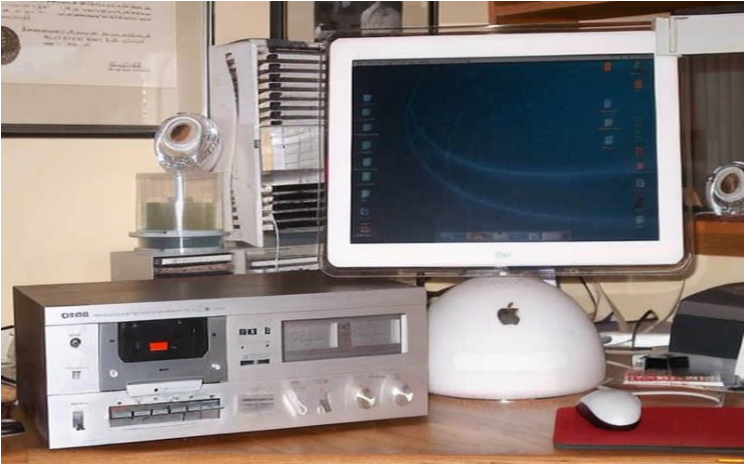}
}

\caption{Qualitative inpainting results on the ImageNet-10 dataset. (a) Original image poisoned with a BadNets trigger flipping ``Cassette Player door'' to ``Tench.'' Subfigures (b--h) illustrate the effect of sequentially removing individual objects or regions until the trigger is detected and purified.}
\label{fig:hig_resolution}
\end{figure*}

\begin{figure*}[!t]
\centering

\subfigure[Original image with physical-style trigger.]{
    \includegraphics[width=0.23\linewidth,height=2.5cm]{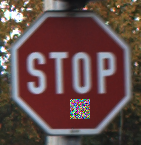}
}
\hfill
\subfigure[Character O removed (Class not changed).]{
    \includegraphics[width=0.23\linewidth,height=2.5cm]{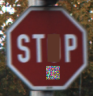}
}
\hfill
\subfigure[Character P removed (Class not changed).]{
    \includegraphics[width=0.23\linewidth,height=2.5cm]{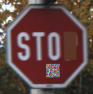}
}
\hfill
\subfigure[Top sign pole removed (Class not changed).]{
    \includegraphics[width=0.23\linewidth,height=2.5cm]{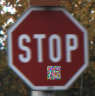}
}

\vspace{2mm}

\subfigure[Character T removed (Class not changed).]{
    \includegraphics[width=0.23\linewidth,height=2.5cm]{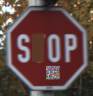}
}
\hfill
\subfigure[Character S removed (Class not changed).]{
    \includegraphics[width=0.23\linewidth,height=2.5cm]{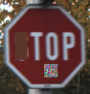}
}
\hfill
\subfigure[Bottom sign pole removed (Class not changed).]{
    \includegraphics[width=0.23\linewidth,height=2.5cm]{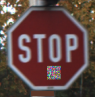}
}
\hfill
\subfigure[Trigger removed (\textbf{Class changed to Stop Sign}).]{
    \includegraphics[width=0.23\linewidth,height=2.5cm]{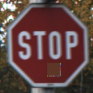}
}

\caption{Qualitative inpainting results for the GTSRB dataset. (a) Original image poisoned with a physical-style trigger flipping ``Stop Sign'' to ``Turn Left.'' Subfigures (b--h) show the effect of sequentially removing individual objects or regions until the trigger is detected and purified.}
\label{fig:low_resolution}
\vspace{7mm}
\end{figure*}

\section{Ablation Study on TRIM’s Detection and Purification Pipeline}

To clearly illustrate how TRIM analyzes, isolates, and removes backdoor triggers, we provide a qualitative ablation study on both high-resolution ImageNet-10 samples and low-resolution GTSRB traffic-sign images. Figure~\ref{fig:hig_resolution} shows an ImageNet-10 example where the input is poisoned with a BadNets trigger (a yellow square in the bottom-right corner), causing the classifier to mislabel the image from ``Cassette Player'' to ``Tench.'' TRIM sequentially removes individual objects or regions (cassette-door, red piece, mouse, paper, monitor base, etc.). As shown in subfigures (b)--(g), removing benign captured regions does \emph{not} change the predicted class, demonstrating that TRIM does not overreact to normal spatial variations. However, once the yellow-square trigger region is removed (Fig.~\ref{fig:hig_resolution}h), the model’s prediction immediately corrects to the true label ``Cassette Player,'' confirming that TRIM successfully isolates the malicious region responsible for misclassification.

Similarly, Figure~\ref{fig:low_resolution} illustrates the step-by-step purification process on a poisoned GTSRB sample, where a physical-style trigger flips a ``Stop Sign'' to ``Turn Left.''  Subfigures (b)--(g) illustrate the removal of alphabetic characters or structural elements of the sign---none of which affect the model’s prediction. Only when the physical-style trigger is removed (Fig.~\ref{fig:low_resolution}h) does the label revert from the attacker-enforced target (``Turn Left'') back to the correct class (“Stop Sign”). This validates that TRIM reliably detects trigger-induced perturbations even under low-resolution, highly compressed settings.

Across both datasets of different resolutions, TRIM (1) accurately localizes subtle, spatially blended, or physically printed triggers, (2) avoids modifying clean or semantically essential regions, and (3) restores correct predictions with minimal visual distortion. These ablations highlight the robustness and precision of TRIM’s region-level detection and purification mechanism.

\begin{figure*}[!t]
    \centering
    \includegraphics[width=0.98\linewidth]{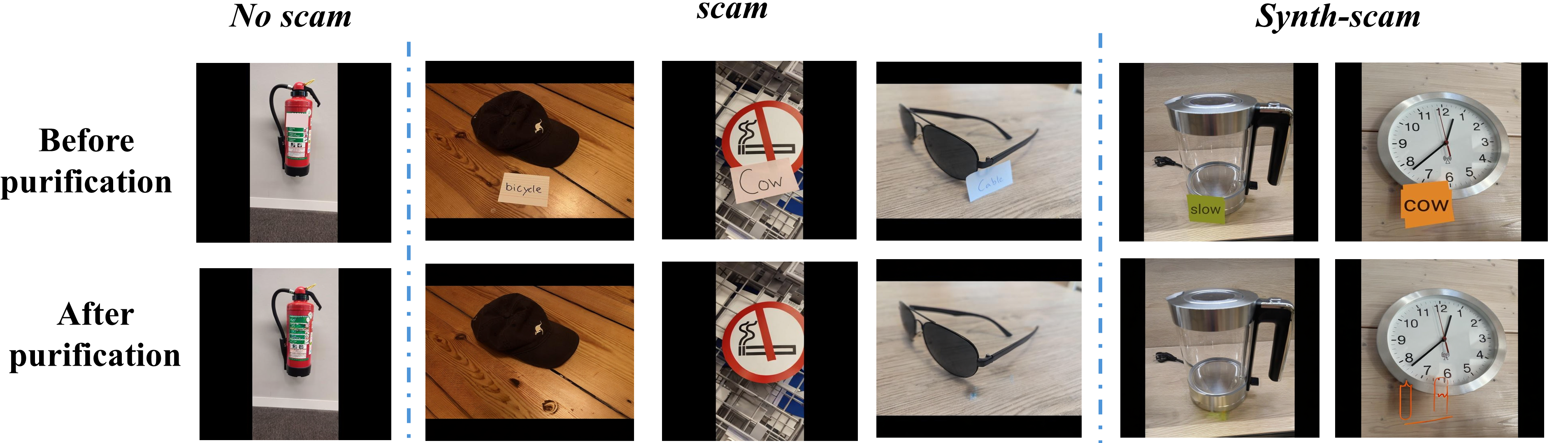}
    \caption{Qualitative evaluation of TRIM against {\em real and synthetic typographic} attacks. The top row ({\em before purification}) shows that adversarial text cues---including  \emph{bicycle}, \emph{cow}, \emph{cable}, and \emph{slow}---successfully mislead modern ViT classifiers and VLMs by overriding the true semantics of fire extinguishers, hats, signs, glasses, kettles, and clocks. Real SCAM attacks (handwritten labels) and Synth-SCAM attacks (digitally created labels) both produce strong misclassification.  The bottom row ({\em after purification}) demonstrates that TRIM reliably removes these malicious triggers while preserving the underlying visual content, restoring correct, object-consistent predictions across all settings.}

    \label{fig:scam_typographic}
\end{figure*}

\section{Evaluating TRIM’s Effectiveness Against Modern Typographic  Attacks} Typographic attacks~\cite{azuma2023defense,gong2025figstep} and the recently introduced SCAM dataset~\cite{westerhoff2025scam} embed adversarial text directly into the visual scene, creating small but highly influential labels. These blended texts exploit the strong textual biases learned by vision transformer (ViT) classifiers and Vision--Language Models (VLMs)~\cite{radford2021learning,li2023blip}, causing predictions to align with the injected text rather than the underlying object. Figure~\ref{fig:scam_typographic} illustrates this vulnerability across three settings: \emph{No Scam} (clean images with a blank sticky note), \emph{SCAM} (handwritten labels), and \emph{Synth-SCAM} (digitally composited labels). Even in the \emph{No Scam} case, the model misclassifies a fire extinguisher as a \emph{car} before purification, highlighting inherent fragility in the visual backbone. In the {\em before-purification row}, models are consistently misled: a fire extinguisher labeled as \emph{car}, hats as \emph{bicycle}, no smoking sign as \emph{cow}, glass as \emph{cable}, and household items as \emph{slow} or \emph{cow}, demonstrating that minimal text cues or learned biases can override robust visual evidence.

After applying TRIM, adversarial text cues are effectively removed while preserving natural image structures. In the \emph{No Scam} case, TRIM restores the correct label \emph{fire extinguisher}, with similar recovery across SCAM and Synth-SCAM examples. The {\em after-purification row} confirms that predictions return to object-consistent labels for all categories, including fire extinguisher, hat, no-smoking sign, glasses, kettle, and clock. These results demonstrate that TRIM generalizes beyond pixel-level triggers, reliably defending against multimodal semantic manipulations. Overall, TRIM provides a unified, model-agnostic defense suitable for both white-box and black-box settings, significantly enhancing the robustness and trustworthiness of modern Vision--language systems.

\end{document}